\documentclass[10pt,journal,compsoc]{IEEEtran}

\ifCLASSOPTIONcompsoc
  \usepackage[nocompress]{cite}
\else
  \usepackage{cite}
\fi

\usepackage[utf8]{inputenc} 
\usepackage[T1]{fontenc}    
\usepackage{hyperref}       
\usepackage{url}            
\usepackage{booktabs}       
\usepackage{amsfonts}       
\usepackage{nicefrac}       
\usepackage{microtype}      
\usepackage{xcolor}         
\usepackage{amsmath}
\usepackage{xspace}
\usepackage{amsfonts}
\usepackage{graphicx}
\usepackage{float}
\usepackage{courier}
\usepackage{wrapfig}
\usepackage{algorithm}
\usepackage{algorithmic}
\usepackage[figuresright]{rotating}
\usepackage{makecell}
\usepackage{bm}
\usepackage{multirow}

\usepackage{hyperref}
\hypersetup{colorlinks=true,linkcolor=blue}

\usepackage{listings}
\usepackage{color}

\newcommand{\mtj}[1]{\textcolor[rgb]{0, 0, 0}{#1}}

\newcommand{\gmh}[1]{\textcolor[rgb]{0, 0, 0}{#1}}

\definecolor{dkgreen}{rgb}{0,0.6,0}
\definecolor{gray}{rgb}{0.5,0.5,0.5}
\definecolor{mauve}{rgb}{0.58,0,0.82}

\graphicspath{{./Images/}, {./figure/Images/}}

\usepackage{silence}
\begin{document}

\title{Attention Mechanisms in Computer Vision:\\ A Survey
}

\author{
        Meng-Hao Guo, Tian-Xing Xu, 
        Jiang-Jiang Liu, Zheng-Ning Liu, Peng-Tao Jiang, 
        Tai-Jiang Mu,
        Song-Hai Zhang,
        Ralph R. Martin, 
        Ming-Ming Cheng,~\IEEEmembership{Senior Member,~IEEE,}
        Shi-Min Hu,~\IEEEmembership{Senior Member,~IEEE,}

\IEEEcompsocitemizethanks{
\IEEEcompsocthanksitem M.H.Guo, T.X.Xu, Z.N.Liu, T.J.Mu, S.H.Zhang and S.M.Hu are with the BNRist, Department of Computer Science and Technology, Tsinghua University, Beijing 100084, China.
\IEEEcompsocthanksitem J.J.Liu, P.T.Jiang and M.M.Cheng are with TKLNDST, College of Computer Science, Nankai University, Tianjin 300350, China.
\IEEEcompsocthanksitem R.R.Martin was with the School of Computer Science and Informatics, Cardiff University, UK.
\IEEEcompsocthanksitem S.M.Hu is the corresponding author.\protect\\ 
E-mail: shimin@tsinghua.edu.cn.}
}

\markboth{Journal of \LaTeX\ Class Files,~Vol.~14, No.~8, August~2015}%
{Shell \MakeLowercase{\textit{et al.}}: Bare Demo of IEEEtran.cls for Computer Society Journals}

\IEEEtitleabstractindextext{%

\IEEEpeerreviewmaketitle

\begin{abstract} 
Humans can naturally and effectively find  
salient regions 
in complex scenes. 
Motivated by this observation, attention mechanisms were introduced into computer vision with the aim of
 imitating this aspect of the human visual system.
Such an attention mechanism can be regarded as a
dynamic weight adjustment process based on  features of the input image.
Attention mechanisms have  achieved 
great success in many visual tasks, including image classification, object detection, semantic segmentation, video understanding, image generation, 3D vision, multi-modal tasks and self-supervised learning.
In this 
survey, we provide a comprehensive review of 
various attention mechanisms in computer vision 
and categorize them according to 
approach,
such as channel attention, spatial attention, 
temporal attention and branch attention;
a {related repository} \url{https://github.com/MenghaoGuo/Awesome-Vision-Attentions} is dedicated to collecting related work.
We also suggest future directions
for attention mechanism research.
\end{abstract}


}

\maketitle

\IEEEdisplaynontitleabstractindextext

\IEEEraisesectionheading{\section{Introduction}\label{sec:introduction}}

%
\IEEEPARstart{M}{ethods} for diverting attention to the most important regions of an image and disregarding irrelevant parts 
are called attention mechanisms; 
the human visual system uses one~\cite{saliency_attention, hayhoe2005eye, rensink2000dynamic, corbetta2002control} 
to assist in analyzing and understanding complex scenes 
efficiently and effectively. 
This in turn has inspired researchers 
to introduce  attention mechanisms 
into computer vision systems to improve their performance.
In a vision system, an attention mechanism 
can be treated as a dynamic selection process 
that is realized by adaptively weighting features 
according to the importance of the input.  
Attention mechanisms have provided benefits
in very many visual tasks, e.g.\ image classification~\cite{senet, woo2018cbam},
object detection~\cite{dai2017deformable, carion2020endtoend_detr}, 
semantic segmentation~\cite{YuanW18_ocnet, Fu_2019_danet}, 
face recognition~\cite{yang2017neural, Wang_2020_CVPR_face}, 
person re-identification~\cite{li2018harmonious, chen2019mixed}, 
action recognition~\cite{Wang_2018_nonlocal, rstan}, 
few-show learning~\cite{Peng_2018_few_shot, he2017single}, 
medical image processing~\cite{oktay2018attention, guan2018diagnose}, 
image generation~\cite{gregor2015draw, zhang2019sagan}, 
pose estimation~\cite{chu2017multicontext}, 
super resolution~\cite{sr_1, zhang2018image},  
3D vision~\cite{Xie_2018_acnn, Guo_2021_pct}, and
multi-modal task~\cite{su2020vlbert, xu2017attngan}.

\begin{figure}[t!]
    \centering
    \includegraphics[width=\linewidth]{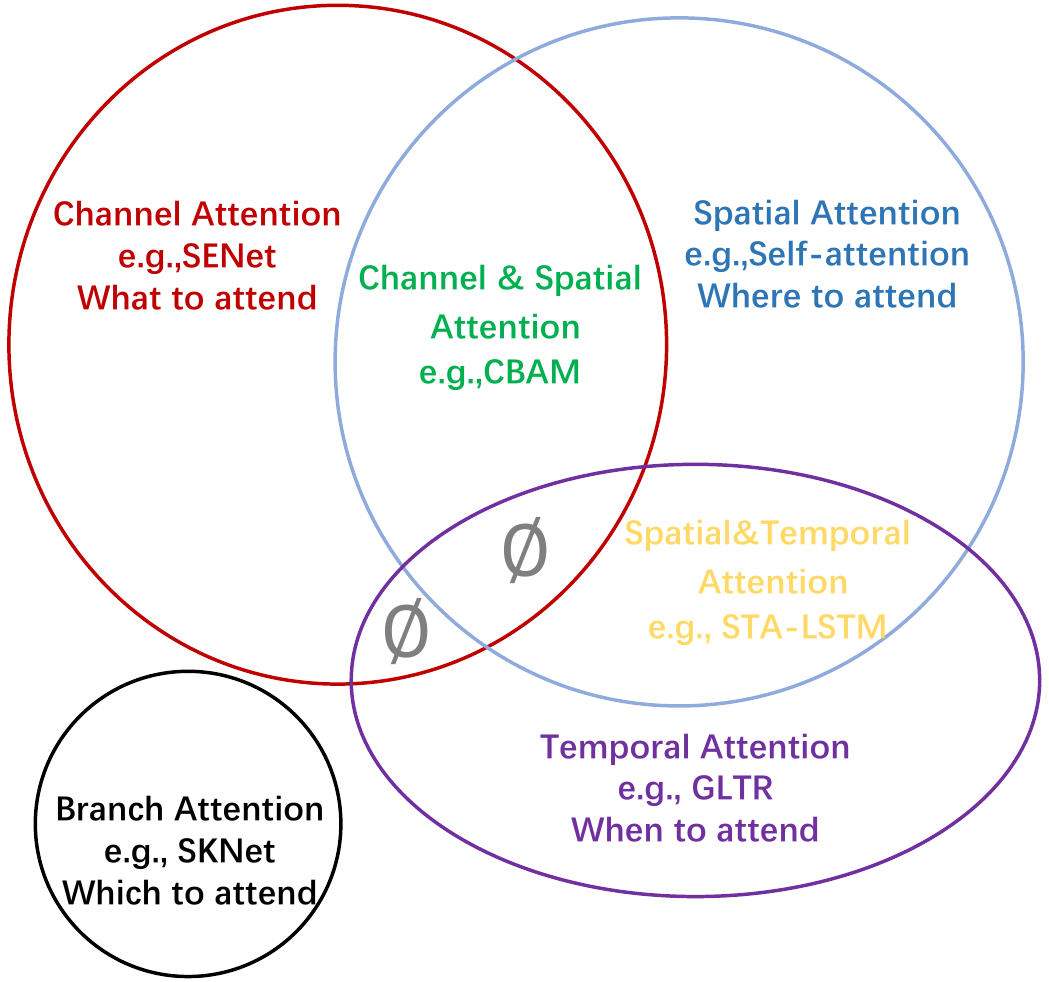}
    \caption{
    Attention mechanisms can be categorised 
    according to data domain. These include four 
    fundamental categories of 
    channel attention, spatial attention, temporal attention 
    and branch attention, 
    and two hybrid categories, combining 
    channel \& spatial attention
    and spatial \& temporal attention. $\emptyset$ means such combinations do not (yet) exist. 
    }
    \label{fig:attention_category}
\end{figure}

\begin{figure}[t]
    \centering
    \includegraphics[width=\linewidth]{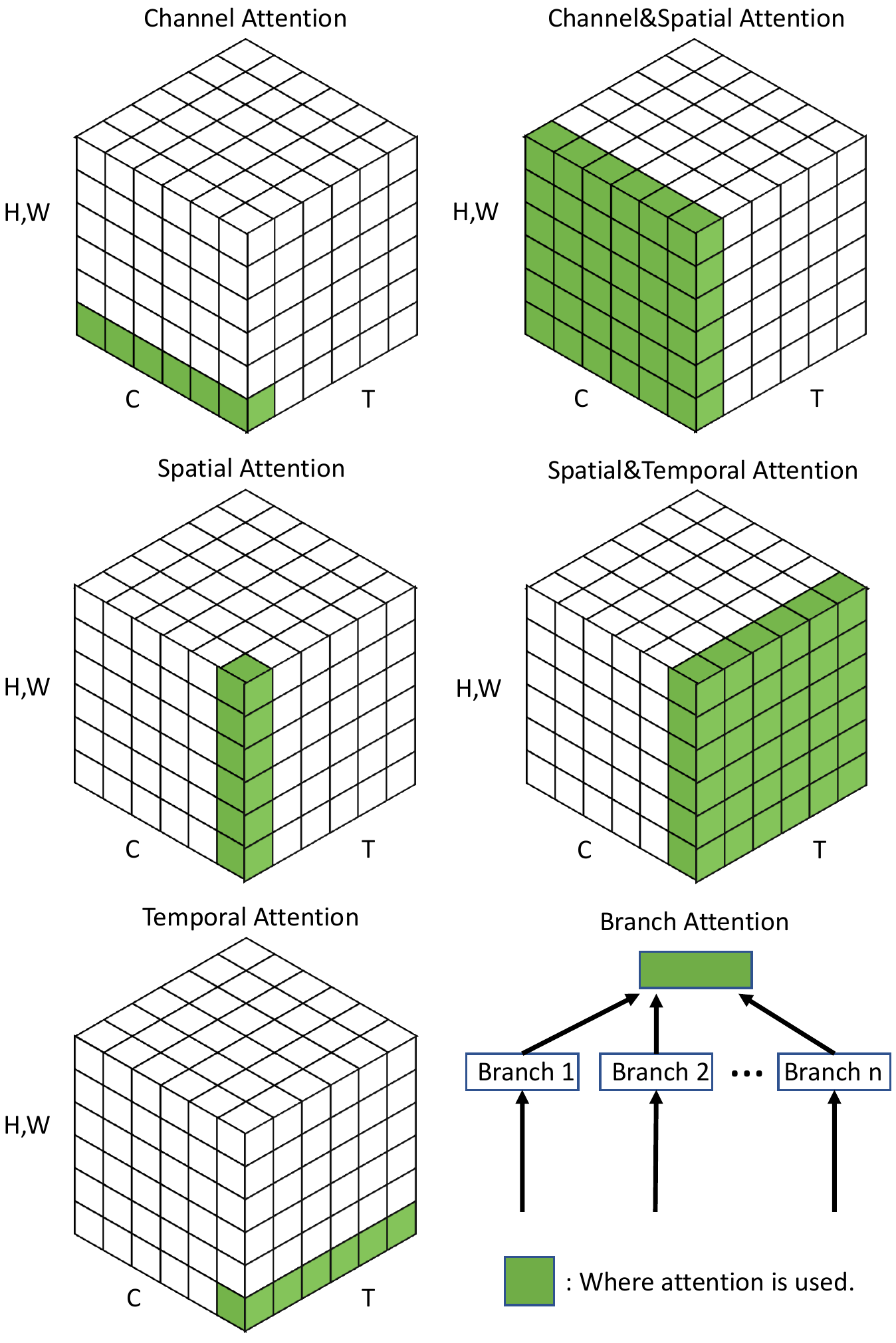}
    \caption{
    Channel, spatial and temporal attention can be regarded as operating on different domains. $C$ represents the channel domain, $H$ and $W$ 
    represent spatial domains, and $T$ means the temporal domain. Branch attention is complementary to these.
    Figure following~\cite{wu2018group}.
    }
    \label{fig:attention_apply_dims}
\end{figure}

\begin{figure*}[t]
    \centering
    \includegraphics[width=\textwidth]{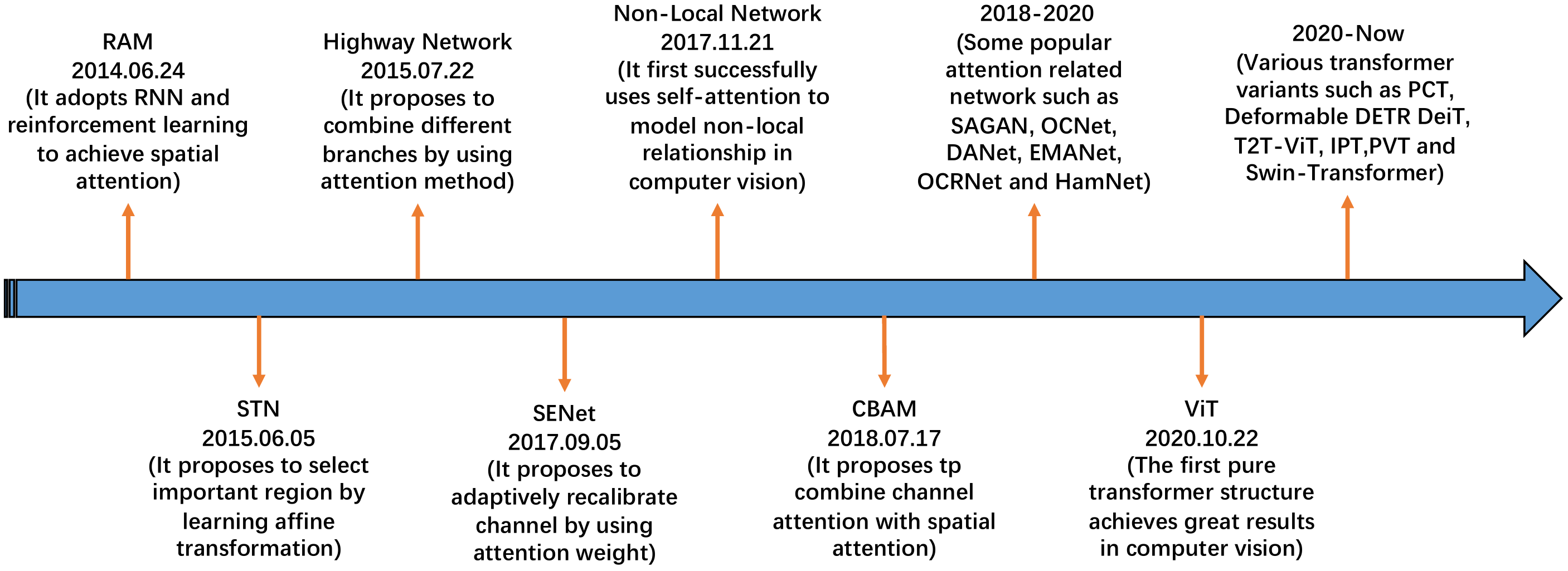}
    \caption{
     Brief summary of key developments in attention 
     in computer vision, which have loosely  occurred in four phases. %
      Phase 1 adopted RNNs 
     to construct attention, a representative method being RAM~\cite{mnih2014recurrent}. 
     %
     Phase 2  explicitly predicted 
     important regions, a representative method being STN~\cite{jaderberg2016spatial}.
     Phase 3 implicitly completed 
     the attention process, a representative method being SENet~\cite{senet}. 
     Phase 4 used
     self-attention methods~\cite{Wang_2018_nonlocal, vaswani2017attention, dosovitskiy2020vit}.
    }
    \label{fig:timeline}
\end{figure*}

\begin{figure*}[t]
    \centering
    \includegraphics[width=\textwidth]{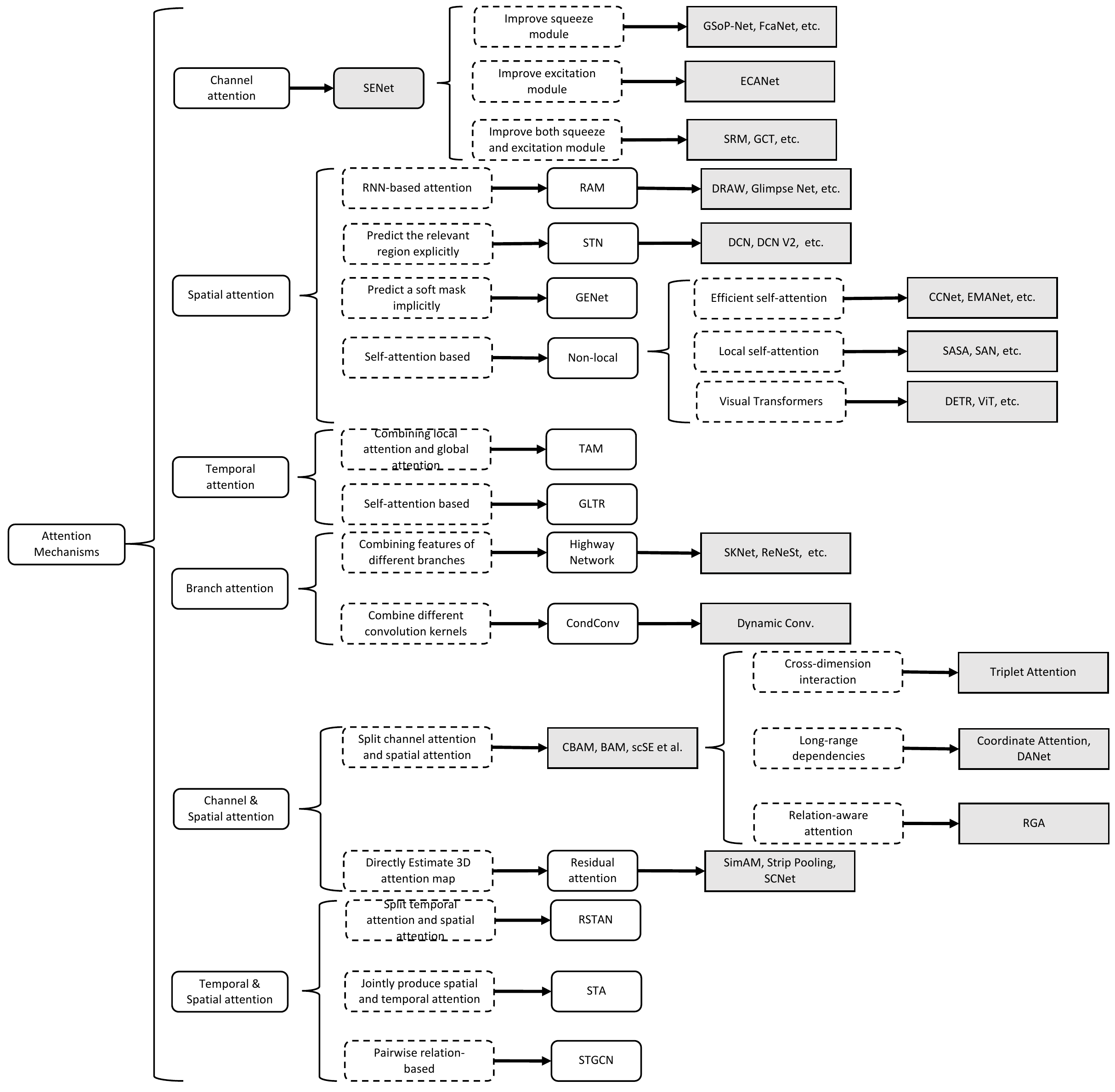}
    \caption{Developmental context of visual attention.
    }
    \label{fig:total_attention_summary}
\end{figure*}

In the past decade, the attention mechanism 
has  played an increasingly important role 
in computer vision; 
Fig.~\ref{fig:timeline}, 
 briefly summarizes the history of 
attention-based models in computer vision 
in the deep learning era. 
Progress can be coarsely divided into four phases.
%
The first phase begins from RAM~\cite{mnih2014recurrent}, 
 pioneering work that combined deep neural networks 
with attention mechanisms. 
It recurrently predicts the important region 
and updates the whole network in an end-to-end manner 
through a policy gradient. 
Later, various works~\cite{gregor2015draw, xu2015show} 
adopted a similar strategy for attention in vision. 
In this phase, recurrent neural networks(RNNs) were necessary tools for an attention mechanism.
At the start of the second phase, 
Jaderberg et al.~\cite{jaderberg2016spatial} proposed the STN 
which introduces a sub-network to predict 
an affine transformation used to to 
select important regions in the input. 
Explicitly predicting discriminatory input features
is the major characteristic of the second phase; 
%
DCNs~\cite{dai2017deformable,Zhu_2019_CVPR_dcnv2} are 
 representative works. 
The third phase began with SENet~\cite{senet} that presented a novel channel-attention network 
which implicitly and adaptively predicts the 
potential key features.
CBAM~\cite{woo2018cbam} and ECANet~\cite{Wang_2020_ECA} 
are representative works of this phase. 
The last phase is the self-attention era. 
Self-attention was firstly proposed in~\cite{vaswani2017attention}
and rapidly provided great advances in the field of 
natural language processing~\cite{vaswani2017attention,devlin2019bert, yang2020xlnet}.
Wang et al.~\cite{Wang_2018_nonlocal} took the lead in 
introducing self-attention to computer vision 
and presented a novel non-local network 
with great success in video understanding 
and object detection. 
It was followed by a series of works 
such as EMANet~\cite{li19ema}, CCNet~\cite{huang2020ccnet},
HamNet~\cite{ham} and the Stand-Alone Network~\cite{ramachandran2019standalone}, which improved speed, quality of results, and generalization capability. 
Recently, various pure deep self-attention networks
(visual transformers)~\cite{dosovitskiy2020vit, yuan2021tokens, wang2021pyramid, liu2021swin, wu2021cvt, Guo_2021_pct, yuan2021volo, dai2021coatnet} have appeared,
showing the huge potential of attention-based models. 
It is clear that attention-based models 
have the potential to replace 
convolutional neural networks and 
become a more powerful \mtj{and} general architecture in computer vision.

The goal of this paper is to summarize and classify  current attention methods in computer vision. Our approach is 
shown in Fig.~\ref{fig:attention_category}
and further explained in Fig.~\ref{fig:attention_apply_dims}:
it is based around data domain. Some methods consider the question of \emph{when} the important data occurs, or others \emph{where} it occurs, etc., and accordingly try to find key times or locations in the data.
We divide existing attention methods 
into six categories which include four basic categories: 
channel attention (\emph{what to pay attention to}~\cite{ChenZXNSLC17}), 
spatial attention (\emph{where to pay attention}), 
temporal attention (\emph{when to pay attention})
and branch channel (\emph{which to pay attention to}), 
along with two hybrid combined categories: 
channel \& spatial attention and spatial \& temporal attention.
These ideas are further briefly summarized together with related works   in Tab.~\ref{Tab_attentions_category}. 

The main contributions of this paper are:
\begin{itemize}
\item  a systematic review of visual attention methods, covering the unified description of attention mechanisms, the development of visual attention mechanisms as well as current research,  
\item  a categorisation grouping attention methods according to their data domain, allowing us to link visual attention methods
independently of their particular application, and
\item suggestions for future research in visual attention.
\end{itemize}

Sec.~\ref{sec:relatedWork} considers 
 related surveys, then 
Sec.~\ref{sec:detailedReview} is the main body of our survey. Suggestions for
future research are given in Sec.~\ref{sec:direction} and finally, we give conclusions in Sec.~\ref{sec:conclusion}.

\begin{table}[!t]
\centering
\caption{Key notation in this paper. Other minor notation is explained where used.}
\label{Tab.notation_defination}
\begin{tabular}{ll}
\hline
\textbf{Symbol} & \textbf{Description} \\
\hline
$X$ & input feature map, $X \in \mathbb{R}^{C\times H\times W}$\\ 
$Y$ & output feature map \\ 
$W$ & learnable kernel weight \\ 
$\text{FC}$ & fully-connected layer \\ 
$\text{Conv}$ & convolution \\ 
$\text{GAP}$ & global average pooling  \\ 
$\text{GMP}$ & global max pooling  \\ 
$[\quad]$ & concatenation  \\ 
$\delta$ & ReLU activation~\cite{relu}  \\ 
$\sigma$ & sigmoid activation  \\ 
$\tanh$ & tanh activation  \\ 
$\text{Softmax}$ & softmax activation  \\ 
$\text{BN}$ & batch normalization~\cite{ioffe2015batch} \\ 
$\text{Expand}$ & expan input by repetition \\ 
\hline
\end{tabular}
\end{table}

\begin{table*}
	\centering
	\caption{Brief summary of attention categories 
	and key related works.}
\label{Tab_attentions_category}
\begin{tabular}{|p{3cm}|p{7cm}|p{6cm}|}
\hline
\textbf{Attention category} & \textbf{Description} & \textbf{Related work} \\
\hline
Channel attention & {Generate attention mask across the channel domain and use it to  select important channels.
} 
& {~\cite{senet, zhang2018encnet, gsopnet, lee2019srm, yang2020gated ,Wang_2020_ECA, qin2021fcanet, diba2019spatiotemporal} ~\cite{chen2019look,shi2020spsequencenet, zhang2018image}} \\
\hline
Spatial attention &  {Generate attention mask across  spatial domains and use it to  select important spatial regions (e.g.~\cite{Wang_2018_nonlocal, hu2019gatherexcite}) or predict the most  relevant spatial position directly (e.g.~\cite{mnih2014recurrent, dai2017deformable})}. &  {~\cite{mnih2014recurrent, gregor2015draw, xu2015show, jaderberg2016spatial, Wang_2018_nonlocal, dosovitskiy2020vit, carion2020endtoend_detr, YuanW18_ocnet} ,~\cite{Xie_2018_acnn, yan2020pointasnl, hu2018relation, Zhang_2019_cfnet, zhang2019sagan, Bello_2019_ICCV_AANet, zhu2019empirical, Li_2020_sprr} ,~\cite{huang2020ccnet, ann, cao2019GCNet, chen2018a2nets, chen2018glore, zhang2019latentgnn, yuan2021segmentation_ocr, yin2020disentangled} ,~\cite{ham, external_attention, ramachandran2019standalone, hu2019lrnet, Zhao_2020_SAN, pmlr-v119-chen20s-igpt, carion2020endtoend_detr, dosovitskiy2020vit} ,~\cite{Guo_2021_pct, chen2021pretrained_ipt, zhao2020pointtransformer, yuan2021tokens, SETR, wang2021pyramid, han2021transformer_tnt, liu2021swin} ,~\cite{liu2021query2label, chen2021empirical_mocov3, bao2021beit, xie2021segformer, hu2019gatherexcite, Zhao_2018_psanet, ba2015multiple, sharma2016action} ,~\cite{girdhar2017attentional, li2016videolstm, yue2018compact,liu2019l2g,paigwar2019attentional,wen2020cf,yang2019modeling, wu2021cvt} ,~\cite{xu2018attention,liu2017end,zheng2018pedestrian, li2018tell,zhang2020relation,xia2019second,zhao2017diversified,zheng2017learning} ,~\cite{fu2017look, anonymous2022dabdetr, yang2021sampling, zheng2019looking, lee2019settransformer, guan2018diagnose} }\\
\hline
Temporal attention & {Generate attention mask in time and use it to  select key frames.} & {~\cite{xu2017jointly,zhang2019scan,chen2018video}} \\
\hline
Branch attention & {Generate attention mask across the different branches and use it to select important branches.} &  {~\cite{srivastava2015training, li2019selective, zhang2020resnest, chen2020dynamic}} \\
\hline
Channel \& spatial attention & {Predict  channel and  spatial attention masks separately (e.g.~\cite{woo2018cbam, park2018bam}) or  generate a joint 3-D  channel, height, width attention mask directly (e.g.~\cite{pmlr-v139-simam, wang2017residual}) and use  it to select important features.} & {~\cite{woo2018cbam, park2018bam, wang2017residual, Liu_2020_scnet, misra2021rotate, ChenZXNSLC17, LinsleySES19},  ~\cite{roy2018recalibrating, Fu_2019_danet, hou2020strip, zhang2020relation, pmlr-v139-simam,you2018pvnet,xie2020mlcvnet} ,~\cite{wang2018mancs,chen2019mixed,chen2019abd, hou2021coordinate, li2018harmonious}} \\
\hline
Spatial \& temporal attention &  {Compute temporal and  spatial attention masks separately (e.g.~\cite{song2016end,rstan}), or produce a joint spatiotemporal attention mask (e.g.~\cite{Fu_2019_STA}), to focus on informative regions.} & {~\cite{gao2019hierarchical,STAT, song2016end, meng2019interpretable, he2021gta,li2018diversity},~\cite{zhang2020multi,shim2020read, liu2021decoupled}} \\
\hline
\end{tabular}
\end{table*}

\section{Other surveys} \label{sec:relatedWork}

In this section, we  briefly compare this paper to various existing surveys which have reviewed attention methods and visual transformers.
Chaudhari et al.~\cite{chaudhari2021attentive} provide a survey of 
attention models in deep neural networks 
which concentrates on their application to natural language processing, while our work focuses on computer vision.
Three more specific surveys~\cite{xu2022transformers, han2021survey, khan2021transformers} summarize the development of visual transformers
while our paper reviews attention mechanisms in vision
more generally,  
not just self-attention mechanisms.
Wang et al.~\cite{wang2016survey} present a survey of attention models in computer vision, but it only considers RNN-based attention models, which form just a part of our survey.
In addition, unlike previous
surveys, 
we provide a classification which groups various attention methods according to their data domain, rather than
according to their field of application.  
Doing so allows us to concentrate on the attention methods in their own right,
rather than treating them as supplementary to other tasks.

\section{Attention methods in computer vision} \label{sec:detailedReview}

\gmh{In this section, we first sum up a general form for the attention mechanism based on the recognition process of human visual system in Sec.~\ref{sec:general}.}
Then we review various categories of attention models given in Fig.~\ref{fig:attention_category}, with a subsection dedicated to each category.
In each, we tabularize representative works for that category.
We also introduce 
that category of attention strategy more deeply,
considering its development
in terms of motivation, formulation and function. 

\subsection{\gmh{General form}} \label{sec:general}
\gmh{
When seeing a scene in our daily life, 
we will focus on the discriminative regions, 
and process these regions quickly. 
The above process can be formulated as:
}

\begin{align}
\label{eq_general}
    Attention = f(g(x), x) 
\end{align}

\gmh{
Here $g(x)$ can represent to generate attention 
which corresponds to 
the process of attending to the discriminative regions. 
$f(g(x),x)$ means processing input $x$ based on the attention $g(x)$ which is consistent with processing critical regions and getting information. 
}

\gmh{
With the above definition, we find that almost all existing attention mechanisms can be written into the above formulation. 
Here we take self-attention~\cite{Wang_2018_nonlocal} and squeeze-and-excitation(SE) attention~\cite{senet} as examples. 
For self-attention, $g(x)$ and $f(g(x), x)$ can be written as
}
\begin{align}
    Q, K, V & = \text{Linear}(x) \\ 
    g(x) & = \text{Softmax}(QK)  \\
    f(g(x), x) & = g(x) V \\
\end{align}

For SE, $g(x)$ and $f(g(x), x)$ can be written as
\begin{align}
    g(x) & = \text{Sigmoid(MLP(GAP}(x)))  \\
    f(g(x), x) & = g(x) x 
\end{align}


In the following, we will introduce various attention mechanisms and specify them to the above formulation.

\subsection{Channel Attention}

In deep neural networks, different channels in different feature maps
usually represent different objects~\cite{ChenZXNSLC17}.
Channel attention adaptively recalibrates the weight of each channel, and  
can be viewed as an object selection process, thus determining \emph{what to pay attention to}. 
Hu et al.~\cite{senet} first proposed the concept of channel attention
and presented SENet for this purpose. 
As Fig.~\ref{fig:total_attention_summary} shows, and we discuss shortly,
three streams of work continue to improve channel attention in different ways.

\begin{figure*}[t]
    \centering
    \includegraphics[width=\textwidth]{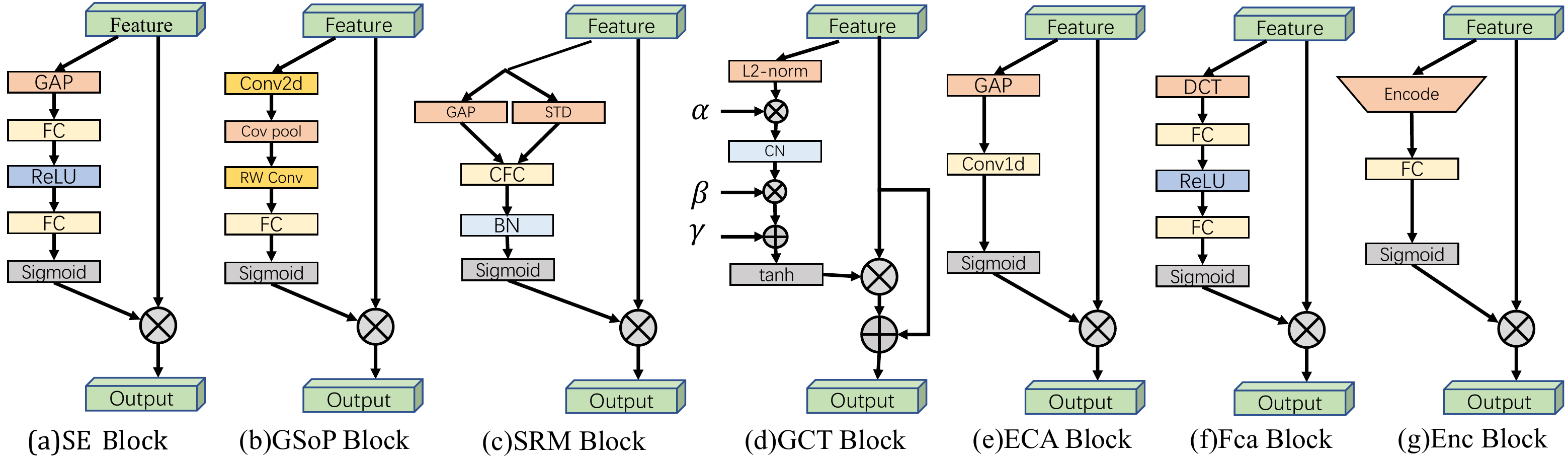}
    \caption{ 
     Various channel attention mechanisms. GAP=global average pooling, GMP=global max pooling, FC=fully-connected layer, Cov pool=Covariance pooling, RW Conv=row-wise convolution, CFC=channel-wise fully connected, CN=channel normalization, DCT=discrete cosine transform.
    }
    \label{fig:channel_attention}
\end{figure*}

In this section, we first 
summarize the representative channel attention works
and specify process $g(x)$ and $f(g(x), x)$ described as Eq.~\ref{eq_general}
in Tab.~\ref{Tab_channel_attentions} and Fig.~\ref{fig:channel_attention}. 
Then we discuss various channel attention methods 
along with their development process respectively.

\subsubsection{SENet}

SENet~\cite{senet}  pioneered channel attention.
 The core  of SENet is a \emph{squeeze-and-excitation} (SE) block which 
 is used to collect global information, capture channel-wise relationships and improve representation ability. 

SE blocks are divided into two parts, a
squeeze module and an excitation module.
Global spatial information is collected in the squeeze module by global average pooling. The excitation module captures channel-wise relationships and outputs an attention vector by using fully-connected layers and non-linear layers (ReLU and sigmoid). Then, each channel of the input feature is scaled by multiplying the corresponding element in the attention vector. Overall, a squeeze-and-excitation block $F_\text{se}$ (with parameter $\theta$) which takes $X$ as input and outputs $Y$ can be formulated 
as:
\begin{align}
\label{eq_se}
    s = F_\text{se}(X, \theta) & = \sigma (W_{2} \delta (W_{1}\text{GAP}(X))) \\
    Y & = s  X
\end{align}
SE blocks play the role of emphasizing important channels while suppressing noise. 
An SE block can be added after each residual unit~\cite{resnet} 
due to their low computational resource requirements. 
However, SE blocks have shortcomings.
In the squeeze module, 
global average pooling is too simple to capture complex global information.
In the excitation module, 
fully-connected layers increase the complexity of the model. 
As  Fig.~\ref{fig:total_attention_summary} indicates,
later works 
attempt to improve the outputs of the squeeze module (e.g.\ GSoP-Net~\cite{gsopnet}), 
reduce the complexity of the model by improving the excitation module (e.g.\ ECANet~\cite{Wang_2020_ECA}),
or improve both the squeeze module and the excitation module 
(e.g.\  SRM~\cite{lee2019srm}).

\subsubsection{GSoP-Net}

An SE block 
captures global information by only using global average pooling (i.e. first-order statistics),  
which limits its modeling capability, in particular the ability to capture high-order statistics.

To address this issue, Gao et al.~\cite{gsopnet} proposed to improve the squeeze module by using a \emph{global second-order pooling} (GSoP) block  to model 
high-order statistics while gathering  global information.

Like an SE block, a GSoP block also has a squeeze module and an excitation module. In the squeeze module,
a GSoP block firstly reduces the number of channels from $c$ to $c'$ ($c' < c$) using a $1 x 1$ convolution, 
then  computes a $c' \times c'$ covariance matrix for the different channels to obtain their correlation. 
Next, row-wise normalization is performed on the covariance matrix. 
Each $(i, j)$ in the normalized covariance matrix explicitly relates channel $i$ to channel $j$. 

In the excitation module,
a GSoP block performs row-wise convolution to 
maintain structural information and output a vector.
Then a fully-connected layer and a sigmoid function are applied 
to get a $c$-dimensional attention vector.
Finally, it multiplies the input features by the attention vector, as in an SE block.
A GSoP block can be formulated as:
\begin{align}
\label{eq_gsop}
    s = F_\text{gsop}(X, \theta) & = \sigma (W \text{RC}(\text{Cov}(\text{Conv}(X)))) \\
    Y & = s  X
\end{align}
Here, $\text{Conv}(\cdot)$ reduces the number of channels,
$\text{Cov}(\cdot)$ computes the covariance matrix and
$\text{RC}(\cdot)$ means row-wise convolution.

By using second-order pooling, GSoP blocks have improved the ability to collect global information over the SE block.
However, this comes at the cost of additional computation.
Thus, a single GSoP block is typically added after several residual blocks.

\subsubsection{SRM}
Motivated by successes in style transfer, 
Lee et al.~\cite{lee2019srm} proposed the lightweight \emph{style-based recalibration module} (SRM).
SRM combines style transfer with an attention mechanism. 
Its main contribution is style pooling
which utilizes both mean and standard deviation of the input features 
to improve its capability to capture global information.
It also adopts a lightweight \emph{channel-wise fully-connected} (CFC) layer, in place of the original fully-connected layer, to
reduce the computational requirements.

Given an input feature map $X \in \mathbb{R}^{C \times H \times W}$, SRM first collects global information by using style pooling ($\text{SP}(\cdot)$) which combines global average pooling and global standard deviation pooling. 
Then a channel-wise fully connected ($\text{CFC}(\cdot)$) layer (i.e.\  fully connected per channel), batch normalization $\text{BN}$ and sigmoid function $\sigma$ are used  to provide the attention vector. Finally,   as in an SE block, the input features are multiplied by the attention vector. Overall, an SRM can be written as:
\begin{align}
\label{eq_srm}
    s = F_\text{srm}(X, \theta) & = \sigma (\text{BN}(\text{CFC}(\text{SP}(X)))) \\
    Y & = s  X
\end{align}
The SRM block 
improves both squeeze and excitation modules, yet can be added after each
residual unit like an SE block.

\subsubsection{GCT}
Due to the computational demand  and number of parameters of the fully connected layer 
in the excitation module, 
it is impractical to use an SE block  after each convolution layer. Furthermore, using fully connected layers to model channel relationships 
  is an implicit procedure.
To overcome the above problems, 
Yang et al.~\cite{yang2020gated} propose the \emph{gated channel transformation} (GCT) 
to efficiently collect information while explicitly modeling channel-wise relationships.  

Unlike previous methods, GCT first collects global information by computing the $l_{2}$-norm of each channel. 
Next, a learnable vector $\alpha$ is applied to scale the feature.
Then a competition mechanism is adopted by 
channel normalization to interact between channels. 
Like other common normalization methods, 
a learnable scale parameter $\gamma$ and bias $\beta$ are applied to 
rescale the normalization.
However, unlike previous methods,
GCT adopts tanh activation to control the attention vector.
Finally, it not only multiplies the input by the attention vector but also adds an identity connection. GCT can be written as: 
\begin{align}
\label{eq_gct}
    s = F_\text{gct}(X, \theta) & = \tanh (\gamma CN(\alpha \text{Norm}(X)) + \beta) \\
    Y & = s  X + X, 
\end{align}
where $\alpha$, $\beta$ and $\gamma$ are trainable parameters. $\text{Norm}(\cdot)$ indicates the $L2$-norm of each channel. $CN$ is  channel normalization.

A GCT block has fewer parameters than an SE block, and as it is  lightweight, 
 can be added after each convolutional layer of a CNN.


\begin{table*}
	\centering
	\caption{Representative channel attention mechanisms ordered by \gmh{category and} publication date. Their key aims are to emphasize important channels and capture global information. Application areas include: Cls = classification, Det = detection, SSeg = semantic segmentation, ISeg = instance segmentation, ST = style transfer, Action = action recognition.\gmh{$g(x)$ and $f(g(x), x)$ are the attention process described by Eq.~\ref{eq_general}. Ranges means the ranges of attention map. S or H means soft or hard attention.(A) channel-wise product. (\uppercase\expandafter{\romannumeral1}) emphasize important channels, (\uppercase\expandafter{\romannumeral2}) capture global information.}}
\label{Tab_channel_attentions}
\begin{tabular}{|p{2cm}|p{2cm}|p{1.5cm}|p{1cm}|p{3cm}|p{1.4cm}|p{1.2cm}|p{0.8cm}|p{1.0cm}|}
\hline
\textbf{Category} & \textbf{Method} & \textbf{Publication} & \textbf{Tasks} & \textbf{$\bm{g(x)}$} & \textbf{$\bm{f(g(x), x)}$} & \textbf{Ranges}  & \textbf{S or H} & \textbf{Goals} \\
\hline
\gmh{Squeeze-and-excitation network} & SENet~\cite{senet} & CVPR2018 & Cls, Det & {global average pooling -> MLP -> sigmoid.}  & (A) & (0,1) & S &  {(\uppercase\expandafter{\romannumeral1}),(\uppercase\expandafter{\romannumeral2})} \\ 
\hline
\gmh{Improve squeeze module} & EncNet~\cite{zhang2018encnet} & CVPR2018 & SSeg & {encoder -> MLP -> sigmoid.} & (A) & (0,1) & S & (\uppercase\expandafter{\romannumeral1}),(\uppercase\expandafter{\romannumeral2}) \\  
\cline{2-9}
 & GSoP-Net~\cite{gsopnet} & CVPR2019 & Cls & {2nd-order pooling -> convolution \& MLP -> sigmoid}  & (A) & (0,1) & S & (\uppercase\expandafter{\romannumeral1}),(\uppercase\expandafter{\romannumeral2}) \\ 
\cline{2-9}
~ & FcaNet~\cite{qin2021fcanet} & ICCV2021 & Cls, Det, ISeg & { discrete cosine transform -> MLP -> sigmoid.} & (A) & (0,1) & S & (\uppercase\expandafter{\romannumeral1}),(\uppercase\expandafter{\romannumeral2}) \\ 
\hline

\gmh{Improve excitation module} & ECANet~\cite{Wang_2020_ECA} & CVPR2020 & Cls, Det, ISeg & { global average pooling -> conv1d -> sigmoid.} & (A) & (0,1) & S & (\uppercase\expandafter{\romannumeral1}),(\uppercase\expandafter{\romannumeral2}) \\ 

\hline

\gmh{Improve both squeeze and excitation module} & SRM~\cite{lee2019srm} & arXiv2019 & Cls, ST & { style pooling -> convolution \& MLP -> sigmoid.}  & (A) & (0,1) & S & (\uppercase\expandafter{\romannumeral1}),(\uppercase\expandafter{\romannumeral2}) \\ 
\cline{2-9}
~ & GCT~\cite{yang2020gated} & CVPR2020 & Cls, Det, Action & { compute $L2$-norm on spatial -> channel normalization -> tanh.}  & (A) & (-1,1) & S & (\uppercase\expandafter{\romannumeral1}),(\uppercase\expandafter{\romannumeral2}) \\ 

\hline
\end{tabular}
\end{table*}

\subsubsection{ECANet}
To avoid high model complexity, SENet reduces the number of channels.
However, this strategy fails to directly model correspondence between weight vectors and inputs,
 reducing the quality of results.
To overcome this drawback, 
Wang et al.~\cite{Wang_2020_ECA} proposed the \emph{efficient channel attention} (ECA) block which instead 
uses a 1D convolution to determine the interaction between channels, instead of dimensionality reduction.

An ECA block has similar formulation to an SE block including a squeeze module for aggregating global spatial information and an efficient excitation module for modeling cross-channel interaction. Instead of indirect correspondence, an ECA block only considers direct interaction between each channel and its $k$-nearest neighbors to control model complexity. Overall, the formulation of an ECA block is:
\begin{align}
\label{eq_eca}
    s = F_\text{eca}(X, \theta) & = \sigma (\text{Conv1D}(\text{GAP}(X))) \\
    Y & = s  X
\end{align}
where $\text{Conv1D}(\cdot)$ denotes 1D convolution with a kernel of shape $k$ across the channel domain, to model local cross-channel interaction. The parameter $k$ decides the coverage of interaction, and in ECA the kernel size $k$ is adaptively determined from the channel dimensionality $C$ instead of by manual tuning, using cross-validation:
\begin{equation}
    k = \psi(C) = \left | \frac{\log_2(C)}{\gamma}+\frac{b}{\gamma}\right |_\text{odd}
\end{equation}
where $\gamma$ and $b$ are hyperparameters. $|x|_\text{odd}$ indicates the nearest odd function of $x$. 

Compared to SENet, ECANet has an 
improved excitation module, and provides an efficient and effective block which can readily be 
 incorporated into various
CNNs.

\subsubsection{FcaNet}

Only using global average pooling in the squeeze module limits  representational ability.
To obtain a more powerful representation ability, 
Qin et al.~\cite{qin2021fcanet} 
rethought global information captured
from the viewpoint of compression 
and analysed global average pooling in the frequency domain. 
They proved that global average pooling
is a special case of the discrete cosine transform (DCT)
and used this observation to propose a novel \emph{multi-spectral channel attention}.

Given an input feature map $X \in \mathbb{R}^{C \times H \times W}$, multi-spectral channel attention first splits $X$ into many parts $x^{i} \in \mathbb{R}^{C' \times H \times W}$. Then it applies a 2D DCT to each part $x^{i}$. Note that a 2D DCT can use pre-processing results to reduce computation. After processing each part,  all results are concatenated into a vector. Finally, fully connected layers, ReLU activation and a sigmoid are used to get the attention vector as in an SE block. This can be formulated as:
\begin{align}
\label{eq_fca}
    s = F_\text{fca}(X, \theta) & = \sigma (W_{2} \delta (W_{1}[(\text{DCT}(\text{Group}(X)))])) \\
    Y & = s  X
\end{align}
where $\text{Group}(\cdot)$ indicates dividing the input into many groups and $\text{DCT}(\cdot)$ is the 2D discrete cosine transform. 

This work based on information compression and discrete cosine transforms achieves excellent performance on the classification task.

\subsubsection{EncNet} 
Inspired by SENet, Zhang et al.~\cite{zhang2018encnet} proposed the \emph{context encoding module} (CEM) incorporating \emph{semantic encoding loss} (SE-loss) to model the relationship between scene context and the probabilities of object categories, thus utilizing global scene contextual information for semantic segmentation.

Given an input feature map $X\in \mathbb R^{C\times H\times W}$, a CEM first learns $K$ cluster centers $D=\{d_1,\dots,d_K\}$ and a set of smoothing factors $S=\{s_1,\dots,s_K\}$ in the training phase. Next, it sums the difference between the local descriptors in the input and the corresponding cluster centers using soft-assignment weights to obtain a permutation-invariant descriptor. Then, it applies aggregation to the descriptors of the $K$ cluster centers instead of concatenation for computational efficiency. Formally, CEM can be written as:
\begin{align}
e_k &= \frac{\sum_{i=1}^N e^{-s_k||X_i-d_k||^2}(X_i-d_k) }{\sum_{j=1}^Ke^{-s_j||X_i-d_j||^2}}\\
e &= \sum_{k=1}^K\phi(e_k) \\
s &= \sigma(W e) \\
Y &= s X
\end{align}
where $d_k \in \mathbb{R}^C$ and $s_k \in \mathbb{R}$ are learnable parameters. $\phi$ denotes batch normalization with ReLU activation. In addition to channel-wise scaling vectors, the compact contextual descriptor $e$ is also applied to compute the SE-loss to regularize training, which improves the segmentation of small objects. 

Not only does CEM enhance class-dependent feature maps, but it also forces the network to consider big and small objects equally by incorporating SE-loss. Due to its lightweight architecture, CEM can be applied to various backbones with only low computational overhead.  

\subsubsection{Bilinear Attention}

Following GSoP-Net~\cite{gsopnet},
Fang et al.~\cite{fang2019bilinear} claimed that previous attention models only use first-order information and disregard higher-order statistical information. They thus proposed a new \emph{bilinear attention block} (bi-attention) to capture  local pairwise feature interactions within each channel, while preserving  spatial information.

Bi-attention employs the \emph{attention-in-attention} (AiA) mechanism to capture  second-order statistical information: the outer point-wise channel attention vectors are computed from the output of the inner channel attention. Formally, given the input feature map $X$, bi-attention first uses  bilinear pooling to capture second-order information
\begin{align}
    \widetilde{x} = \text{Bi}(\phi(X)) =  \text{Vec}(\text{UTri}(\phi(X) \phi(X)^T))
\end{align}
where $\phi$ denotes an embedding function used for dimensionality reduction, $\phi(x)^T$ is the transpose of $\phi(x)$ across the channel domain, $\text{Utri}(\cdot)$ extracts the upper triangular elements of a matrix and $\text{Vec}(\cdot)$ is vectorization. Then bi-attention applies the inner channel attention mechanism to the feature map $\widetilde{x}\in\mathbb{R}^{\frac{c'(c'+1)}{2}\times H\times W}$
\begin{align}
    \widehat{x} &=  \omega(\text{GAP}(\widetilde{x})) \varphi(\widetilde{x})
\end{align}
Here $\omega$ and $\varphi$ are embedding functions. Finally the output feature map $\widehat{x}$ is used to compute the spatial channel attention weights of the outer point-wise attention mechanism:
\begin{align}
    s &= \sigma(\widehat{x})\\
    Y &= s X 
\end{align}
The bi-attention block uses bilinear pooling to model the local pairwise feature interactions along each channel, while preserving the spatial information. Using the proposed AiA, the model pays more attention to higher-order statistical information compared with other attention-based models. Bi-attention can be incorporated into 
any CNN backbone to improve its representational power while suppressing noise.

\subsection{Spatial Attention}

Spatial attention can be seen as an adaptive spatial region selection 
mechanism: \emph{where to pay attention}. 
As  Fig.~\ref{fig:total_attention_summary} shows,
RAM~\cite{mnih2014recurrent}, STN~\cite{jaderberg2016spatial},
GENet~\cite{hu2019gatherexcite} and Non-Local~\cite{Wang_2018_nonlocal}
are representative of different kinds of spatial attention methods. 
RAM represents  RNN-based methods. 
STN represents those use 
a sub-network to explicitly predict relevant regions.
GENet represents 
those that use a sub-network implicitly to predict a soft mask
to select  important regions.
Non-Local represents self-attention related methods.
In this subsection, we first 
summarize representative spatial attention mechanisms
and specify process $g(x)$ and $f(g(x), x)$ described as Eq.~\ref{eq_general}
in Tab.~\ref{Tab_spatial_attentions}, then  discuss them according to Fig.~\ref{fig:total_attention_summary}.

\begin{table*}
	\centering
	\caption{Representative spatial attention mechanisms sorted by \gmh{category and }date. Application areas include: Cls = classification, FGCls = fine-grained classification, Det = detection, SSeg = semantic segmentation, ISeg = instance segmentation, ST = style transfer, Action = action recognition, ICap = image captioning.\gmh{$g(x)$ and $f(g(x), x)$ are the attention process described by Eq.~\ref{eq_general}. Ranges means the ranges of attention map. S or H means soft or hard attention.(A) choose region according to the prediction. (B) element-wise product, (C)aggregate information via attention map. (\uppercase\expandafter{\romannumeral1})focus the network on  discriminative regions, (\uppercase\expandafter{\romannumeral2}) avoid excessive computation  for large input images, (\uppercase\expandafter{\romannumeral3}) provide more transformation invariance, (\uppercase\expandafter{\romannumeral4}) capture long-range dependencies, (\uppercase\expandafter{\romannumeral5}) denoise input feature map (\uppercase\expandafter{\romannumeral6}) adaptively aggregate neighborhood  information, (\uppercase\expandafter{\romannumeral7}) reduce inductive bias. }}
\label{Tab_spatial_attentions}
\begin{tabular}{|p{2cm}|p{2cm}|p{1.5cm}|p{1cm}|p{3cm}|p{1.4cm}|p{1.2cm}|p{0.8cm}|p{1.0cm}|}
\hline
\textbf{Category} & \textbf{Method} & \textbf{Publication} & \textbf{Tasks} & \textbf{$\bm{g(x)}$} & \textbf{$\bm{f(g(x), x)}$} & \textbf{Ranges}  & \textbf{S or H} & \textbf{Goals} \\
\hline
\gmh{RNN-based methods} & RAM~\cite{mnih2014recurrent} & NIPS2014 & Cls &{use RNN to recurrently predict  important regions}  & (A) & {(0,1)} & H & {(\uppercase\expandafter{\romannumeral1}), (\uppercase\expandafter{\romannumeral2}).} \\ 
\cline{2-9}
& Hard and soft attention~\cite{xu2015show} & ICML2015 & ICap &{a)compute similarity between visual features and  previous hidden state -> interpret attention weight.}  & (C) & (0,1) & S, H & {(\uppercase\expandafter{\romannumeral1}).} \\ 
\hline
\gmh{Predict the relevant region explictly} & STN~\cite{jaderberg2016spatial} & NIPS2015 & Cls, FGCls &{ use sub-network to predict an affine transformation.} & (A) & {(0,1)} & H & {(\uppercase\expandafter{\romannumeral1}), (\uppercase\expandafter{\romannumeral3}).} \\ 
\cline{2-9}

& DCN~\cite{dai2017deformable} & ICCV2017 & Det, SSeg &{use sub-network to predict offset coordinates.} & (A) & {(0,1)} & H & {(\uppercase\expandafter{\romannumeral1}), (\uppercase\expandafter{\romannumeral3}).} \\ 
\hline
\gmh{Predict the relevant region implictly}  & GENet~\cite{hu2019gatherexcite} & NIPS2018 & Cls, Det &{average pooling or depth-wise convolution -> interpolation ->  sigmoid }  & (B) & (0,1) & S & {(\uppercase\expandafter{\romannumeral1}).} \\ 
\cline{2-9}
&  PSANet~\cite{Zhao_2018_psanet} & ECCV2018 & SSeg &{predict an attention map using a sub-network.}  & (C) & (0,1) & S &  {(\uppercase\expandafter{\romannumeral1}), (\uppercase\expandafter{\romannumeral4}).} \\ 

\hline
\gmh{Self-attention based methods}& Non-Local~\cite{Wang_2018_nonlocal} & CVPR2018 & Action, Det, ISeg &{Dot product between query and key -> softmax   }  & (C) & (0,1) & S &  {(\uppercase\expandafter{\romannumeral1}), (\uppercase\expandafter{\romannumeral4}), (\uppercase\expandafter{\romannumeral5})}  \\  
\cline{2-9}
& SASA~\cite{ramachandran2019standalone} & NeurIPS2019 & Cls, Det &{Dot product between query and key -> softmax.}  & (C) & (0,1) & S &  {(\uppercase\expandafter{\romannumeral1}), (\uppercase\expandafter{\romannumeral6})}  \\  
\cline{2-9}
& ViT~\cite{dosovitskiy2020vit} & ICLR2021 & Cls &{divide the feature map into multiple groups -> Dot product between query and key -> softmax.}  & (C) & (0,1) & S & {(\uppercase\expandafter{\romannumeral1}),(\uppercase\expandafter{\romannumeral4}), (\uppercase\expandafter{\romannumeral7}).  } \\ 
\hline
\end{tabular}
\end{table*}

\subsubsection{RAM} 
Convolutional neural networks have huge computational costs, especially
for large inputs. 
In order to concentrate limited computing resources on important regions, 
Mnih et al.~\cite{mnih2014recurrent} proposed the \emph{recurrent attention model} (RAM) that
adopts RNNs~\cite{HochSchm97_lstm} 
and reinforcement learning (RL)~\cite{sutton2000policy} to 
make the network learn  where to pay attention. 
RAM  pioneered the use of  RNNs for visual attention, and was followed
by many other RNN-based methods~\cite{gregor2015draw, xu2015show, ba2015multiple}. 

\begin{figure*}[t]
    \centering
    \includegraphics[width=\textwidth]{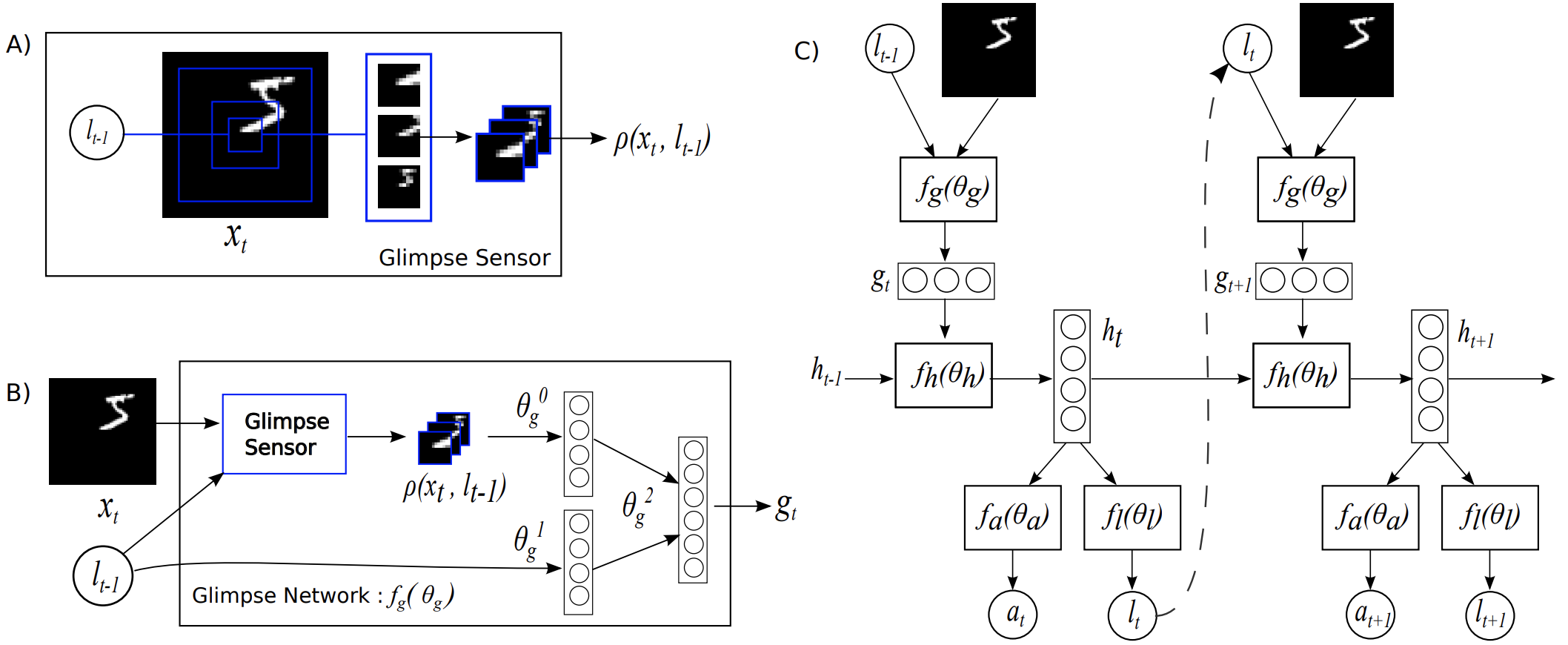}
    \caption{Attention process in RAM~\cite{mnih2014recurrent}. (A): a glimpse sensor takes image and center coordinates as input and outputs multiple resolution patches. (B): a glimpse network includes a glimpse sensor, taking  image and center coordinates as input and outputting a feature vector. (C) the entire network recurrently uses a glimpse network, outputting the predicted result as well as the next center coordinates. Figure is taken from ~\cite{mnih2014recurrent}.
    }
    \label{fig:ram}
\end{figure*}

As shown in Fig.~\ref{fig:ram}, the RAM has three key elements: (A) a glimpse sensor, (B) a glimpse network and (C) an RNN model. The glimpse sensor takes a coordinate $l_{t-1}$ and an image $X_{t}$. It outputs multiple resolution patches $\rho(X_{t}, l_{t-1})$ centered on $l_{t-1}$.  The glimpse network $f_{g}(\theta(g))$ includes a glimpse sensor and outputs the feature representation $g_{t}$ for input coordinate $l_{t-1}$ and image $X_{t}$. The RNN model considers $g_{t}$ and an internal state $h_{t-1}$ and outputs the next center coordinate $l_{t}$ and the action $a_{t}$, e.g.\ the softmax result in an image classification task. Since the whole process is not differentiable, it applies reinforcement learning strategies in the update process. 

This provides a simple but effective method to focus the network on key regions, thus reducing the number of calculations performed by the network,
\gmh{especially for large inputs,
}
while improving image classification results.

\subsubsection{Glimpse Network}
Inspired by how humans perform visual recognition sequentially, Ba et al.~\cite{ba2015multiple} proposed a deep recurrent network, similar to RAM~\cite{mnih2014recurrent}, capable of processing a multi-resolution crop of the input image, called a glimpse, for multiple object recognition task. The proposed network updates its hidden state using a glimpse as input, and then predicts a new object as well as the next glimpse location at each step. The glimpse is usually much smaller than the whole image, which makes the network computationally efficient. 

The proposed deep recurrent visual attention model consists of a context network,
glimpse network, recurrent network, emission network, and classification network. 
First, the context network takes the down-sampled whole image 
as input to provide the initial state for the recurrent network 
as well as the location of the first glimpse.
Then, at the current time step $t$, 
given the current glimpse $x_t$ and its location tuple $l_t$, 
the goal of the glimpse network is to extract useful information,
expressed as
\begin{align}
    g_t = f_\text{image}(X) \cdot f_\text{loc}(l_t)
\end{align}
where $f_\text{image}(X)$ and $f_\text{loc}(l_t)$ are non-linear functions which both  output vectors having the same dimension, and $\cdot$ denotes  element-wise product, used for fusing information from two branches. Then, the recurrent network, which consists of two stacked recurrent layers, aggregates information gathered from each individual glimpse. The outputs of the recurrent layers are:
\begin{align}
    r_t^{(1)} &= f_\text{rec}^{(1)}(g_t, r_{t-1}^{(1)}) \\
    r_t^{(2)} &= f_\text{rec}^{(2)}(r_t^{(1)}, r_{t-1}^{(2)})
\end{align}
Given the current hidden state $r_t^{(2)}$ of the recurrent network, 
the emission network predicts where to crop the next glimpse.
Formally, it can be written as
\begin{align}
    l_{t+1} = f_\text{emis}(r_t^{(2)})
\end{align}
Finally, the classification network outputs a prediction for the class label $y$ based on the hidden state $r_{t}^{(1)}$ of the recurrent network
\begin{align}
    y = f_\text{cls}(r_t^{(1)})
\end{align}

Compared to a CNN operating on the entire image,
the computational cost of the proposed model is much lower,
and it can naturally tackle images of different sizes
because it only processes a glimpse in each step.
Robustness is additionally improved by the recurrent attention mechanism,
which also alleviates the problem of over-fitting. 
This pipeline can be incorporated into any state-of-the-art CNN backbones or RNN units.

\subsubsection{Hard and soft attention}

To visualize where and what an image caption generation model should focus on, Xu et al.~\cite{xu2015show} introduced an attention-based model as well as two variant attention mechanisms,  \emph{hard attention} and \emph{soft attention}.

Given a set of feature vectors $\bm{a} = \{a_1,\dots,a_L\}, a_i\in \mathbb{R}^D$ extracted from the input image, the model aims to produce a caption by generating one word at each time step. Thus they adopt a long short-term memory (LSTM) network as a decoder; an attention mechanism is used to generate a contextual vector $z_t$ conditioned on the feature set $\bm{a}$ and the previous hidden state $h_{t-1}$, where $t$ denotes the time step. Formally, the weight $\alpha_{t,i}$ of the feature vector $a_i$ at the $t$-th time step is defined as
\begin{align}
    e_{t,i} &= f_\text{att} (a_i, h_{t-1})\\
    \alpha_{t,i} &= \frac{\exp(e_{t,i})}{\sum_{k=1}^L \exp(e_{t,k})}
\end{align}
where $f_\text{att}$ is implemented by a multilayer perceptron conditioned on the previous hidden state $h_{t-1}$. The positive weight $\alpha_{t,i}$ can be interpreted either as the probability that location $i$ is the right place to focus on (hard attention), or as the relative importance of location $i$ to the next word (soft attention). To obtain the contextual vector $z_t$, the hard attention mechanism assigns a multinoulli distribution parametrized by $\{\alpha_{t,i}\}$ and views $z_t$ as a random variable:
\begin{align}
    p(s_{t,i} &= 1 | \bm{a}, h_{t-1}) = \alpha_{t,i} \\
    z_t &= \sum_{i=1}^L s_{t,i}a_i
\end{align}

On the other hand, the soft attention mechanism directly uses the expectation of the context vector $z_t$, 
\begin{align}
    z_t &= \sum_{i=1}^L \alpha_{t,i}a_i
\end{align}

The use of the attention mechanism improves the interpretability
of the image caption generation process 
by allowing the user to understand what and where the model is focusing on. 
It also helps to improve the representational capability of the network.

\subsubsection{Attention Gate}

Previous approaches to MR segmentation 
usually operate on particular regions of interest (ROI),
which requires excessive and wasteful use of computational resources and model parameters. To address this issue, Oktay et al. ~\cite{oktay2018attention} proposed a simple and yet effective mechanism, the \emph{attention gate} (AG), to focus on targeted regions while suppressing feature activations in irrelevant regions.

Given the input feature map $X$ and the gating signal $G\in \mathbb{R}^{C'\times H\times W}$ which is collected at a coarse scale and contains contextual information, the attention gate uses additive attention to obtain the gating coefficient. Both the input $X$ and the gating signal are first linearly mapped to an $\mathbb{R}^{F\times H\times W}$ dimensional space, and then the output is squeezed in the channel domain to produce a spatial attention weight map $ S \in \mathbb{R}^{1\times H\times W}$. The overall process can be written as
\begin{align}
    S &= \sigma(\varphi(\delta(\phi_x(X)+\phi_g(G)))) \\
    Y &= S X
\end{align}
where $\varphi$, $\phi_x$ and $\phi_g$ are linear transformations implemented as $1\times 1$ convolutions. 

The attention gate guides the model's attention to important regions while suppressing feature activation in unrelated areas. It substantially enhances the representational power of the model without a significant increase in computing cost or number of model parameters due to its lightweight design. It is general and modular, making it simple to use in various CNN models.

\subsubsection{STN}
The property of translation equivariance makes CNNs suitable for processing image data. 
However, CNNs lack other transformation invariance such as
rotational invariance, scaling invariance and warping invariance.
To achieve these attributes while making CNNs focus on  important regions,
Jaderberg et al.~\cite{jaderberg2016spatial} proposed \emph{spatial transformer networks} (STN) that 
use an explicit procedure to learn invariance to translation, scaling, rotation and other more general warps, 
making the network pay attention to the most relevant regions.  
STN was the first attention mechanism to explicitly 
predict  important regions 
and provide a deep neural network with transformation invariance.
Various following works~\cite{dai2017deformable, Zhu_2019_CVPR_dcnv2} have had even greater success. 

Taking a 2D image as an example, a 2D affine transformation can be formulated as: 
\begin{align}
\label{eq_2d_affine}
    \begin{bmatrix} \theta_{11} & \theta_{12} & \theta_{13} \\ \theta_{21} & \theta_{22} & \theta_{23} \end{bmatrix} &= f_\text{loc}(U) \\ 
    \begin{pmatrix} x_{i}^{s} \\ y_{i}^{s} \end{pmatrix} &= \begin{bmatrix} \theta_{11} & \theta_{12} & \theta_{13} \\ \theta_{21} & \theta_{22} & \theta_{23} \end{bmatrix}  \begin{pmatrix} x_{i}^{t} \\ y_{i}^{t} \\ 1 \end{pmatrix}.
\end{align}
Here, $U$ is the input feature map, and $f_\text{loc}$ can be any differentiable function, such as a lightweight fully-connected network or convolutional neural network. $x_{i}^{s}$ and $y_{i}^{s}$ are  coordinates in the output feature map, while $x_{i}^{t}$ and $y_{i}^{t}$ are  corresponding coordinates in the input feature map and the $\theta$ matrix is the learnable affine matrix. After obtaining the correspondence, the network can sample relevant input regions using the correspondence. 
To ensure that the whole process is differentiable and can be updated in an end-to-end manner,  bilinear sampling is used to sample the input features 

STNs focus on discriminative regions automatically
and  learn invariance to some geometric transformations.

\subsubsection{Deformable Convolutional Networks}
With similar 
purpose to 
STNs, Dai et al.~\cite{dai2017deformable} proposed \emph{deformable convolutional networks} (deformable ConvNets) 
to be invariant to geometric
transformations, but they pay attention to the important regions in a different manner. 

Specifically, deformable ConvNets do not learn an affine transformation. They divide convolution into two steps,
firstly sampling features on a regular grid $\mathcal{R}$ from the input feature map, then aggregating sampled features by weighted summation using a convolution kernel. The process can be written as:
\begin{align}
    Y(p_{0}) &= \sum_{p_i \in \mathcal{R}} w(p_{i}) X(p_{0} + p_{i}) \\
    \mathcal{R}  &= \{(-1,-1), (-1, 0), \dots, (1, 1)\}
\end{align}
The deformable convolution augments the sampling process by introducing a group of learnable offsets $\Delta p_{i}$ which can be generated by a lightweight CNN. Using the offsets $\Delta p_{i}$, the deformable convolution can be formulated as:
\begin{align}
    Y(p_{0}) &= \sum_{p_i \in \mathcal{R}} w(p_{i}) X(p_{0} + p_{i} + \Delta p_{i}). 
\end{align}
Through the above method, adaptive sampling is achieved.
However, $\Delta p_{i}$ is a floating point value
unsuited to grid sampling. 
To \gmh{address} this problem, bilinear interpolation is used. Deformable RoI pooling is also used, which greatly improves object detection. 

Deformable ConvNets adaptively select \gmh{the} important regions and enlarge the valid receptive field of convolutional neural networks; this is important in object detection and semantic segmentation tasks.

\subsubsection{Self-attention and variants}

Self-attention was proposed and has had  great success in the field of \emph{natural language processing} (NLP)~\cite{bahdanau2016neural, vaswani2017attention, lin2017structured, devlin2019bert, yang2020xlnet, dai2019transformerxl, choromanski2021rethinking}. Recently, it has also shown the potential to become a 
dominant tool in computer vision~\cite{Wang_2018_nonlocal, dosovitskiy2020vit, carion2020endtoend_detr, pmlr-v119-chen20s-igpt, zhu2021deformable}. Typically,
self-attention is used as a spatial attention mechanism to capture  global information. We now summarize \gmh{the} self-attention mechanism and its common variants in computer vision.

Due to the localisation of the convolutional operation, CNNs have inherently narrow receptive fields~\cite{liu2015parsenet,Peng_2017_large_kernel}, 
which limits the ability of CNNs to understand scenes globally.
To increase the receptive field, Wang et al.~\cite{Wang_2018_nonlocal}  introduced self-attention into computer vision. 

Taking a 2D image as an example, given a feature map $F \in \mathbb{R}^{C \times H \times W}$, self-attention first computes the queries, keys and values $Q, K, V \in \mathbb{R}^{C' \times N}, N = H \times W$ by linear projection and reshaping operations. Then self-attention can be formulated as: 
\begin{align}
\label{eq_self_attn}
    A = (a)_{i,j} & = \text{Softmax}(Q  K^T), \\
    Y & = A V, 
\end{align}
where $A \in \mathbb{R}^{N \times N} $ is the attention matrix and $\alpha_{i,j}$ is the relationship between the $i$-th and $j$-th elements. The whole process is shown in Fig.~\ref{fig:vit}(left).  Self-attention is a powerful tool to model global information and is useful in many visual tasks~\cite{YuanW18_ocnet, Xie_2018_acnn, yan2020pointasnl, hu2018relation, Zhang_2019_cfnet, zhang2019sagan, Bello_2019_ICCV_AANet, zhu2019empirical,Li_2020_sprr}. 

However, the  self-attention mechanism has several shortcomings, 
particularly its quadratic complexity, 
which limit its applicability.
Several
variants have been introduced to
alleviate these problems.
The \emph{disentangled non-local} approach~\cite{yin2020disentangled} improves self-attention's accuracy and effectiveness, but most variants focus on reducing its computational complexity.

CCNet~\cite{huang2020ccnet} regards the self-attention operation as a graph convolution and 
replaces the densely-connected graph processed by self-attention with several sparsely-connected graphs. To do so, it proposes \emph{criss-cross attention} which considers row attention and column attention recurrently to obtain global information. 
CCNet reduces the complexity of self-attention from $O(N^{2})$ to $O(N \sqrt{N})$. 

EMANet~\cite{li19ema} views self-attention in terms of expectation maximization (EM). It proposes \emph{EM attention} which adopts the EM algorithm to get a set of compact bases  instead of using all points as  reconstruction bases. This reduces the complexity from $O(N^{2})$ to $O(NK)$, where $K$ is the number of compact bases.

ANN~\cite{ann} suggests that using all positional features as key and vectors is redundant and adopts spatial pyramid pooling~\cite{zhao2017pspnet, He_2014_spp} to obtain a few representative key and value features to use instead, to reduce computation.

GCNet~\cite{cao2019GCNet} analyses the attention map used in self-attention and finds that the global contexts obtained by self-attention are similar for different query positions in the same image. Thus, it first proposes to predict a single attention map shared by all query points, and then gets global information from a weighted sum of input features according to this attention map. This is like average pooling, but  is a more general process for collecting global information. 

$A^{2}$Net~\cite{chen2018a2nets} is motivated by SENet to divide attention into feature gathering and feature distribution processes,  using two different kinds of attention. The first  aggregates global information via second-order attention pooling and the second distributes the global descriptors by soft selection attention.

GloRe~\cite{chen2018glore} understands self-attention from a graph learning perspective. It first collects $N$ input features into $M  \ll N$ nodes and then learns an adjacency matrix of global interactions between nodes. Finally, the nodes distribute global information to input features. A similar idea can be found in LatentGNN~\cite{zhang2019latentgnn}, MLP-Mixer~\cite{tolstikhin2021mlpmixer} and ResMLP~\cite{touvron2021resmlp}. 

OCRNet~\cite{yuan2021segmentation_ocr} proposes the concept of \emph{object-contextual representation} which is a weighted aggregation of all object regions' representations in the same category, such as a weighted average of all car region representations. It replaces the key and vector with this object-contextual representation leading to successful improvements in both  speed and effectiveness. 

The \emph{disentangled non-local} approach was motivated by~\cite{cao2019GCNet, Wang_2018_nonlocal}. Yin et al~\cite{yin2020disentangled} deeply analyzed the self-attention mechanism resulting in the core idea of decoupling self-attention into a pairwise term and a unary term. The pairwise term focuses on modeling relationships while the unary term focuses on salient boundaries. This decomposition prevents unwanted interactions between the two terms, greatly  improving semantic segmentation, object detection and action recognition.

HamNet~\cite{ham} models capturing global relationships as a low-rank completion problem and designs a series of white-box methods to capture global context using matrix decomposition. This not only reduces the complexity, but increases the interpretability of self-attention.

EANet~\cite{external_attention} proposes that self-attention should only consider correlation in a single sample and should ignore potential relationships between different samples. To explore the correlation between different samples and reduce computation, it makes use of an external attention that adopts learnable, lightweight and shared key and value vectors. It further reveals that using softmax to normalize the attention map is not optimal and presents double normalization as a better alternative.

In addition to being a complementary approach to CNNs, self-attention also can be used to replace convolution operations for aggregating neighborhood information. Convolution operations can be formulated as dot products between the input feature $X$ and a convolution kernel $W$:
\begin{equation}
\label{eq_conv}
    Y_{i,j}^{c}  = \sum_{a,b \in \{0, \dots, k-1\}} W_{a,b,c} X_{\hat{a}, \hat{b}} 
    \end{equation}
where 
\begin{equation}
  \hat{a}  = i+a-\lfloor k/2 \rfloor, \qquad
    \hat{b}  = j+b-\lfloor k/2 \rfloor,
    \end{equation}
$k$ is the kernel size and $c$ indicates the channel. 
The above formulation can be viewed as a process of aggregating neighborhood information by using a weighted sum through a convolution kernel. The process of aggregating neighborhood information can be defined more generally as:
\begin{align}
\label{eq_general_conv}
    Y_{i,j} &=  \sum_{a,b \in \{0, \dots, k-1\}} \text{Rel}(i,j,\hat{a},\hat{b}) f(X_{\hat{a},\hat{b}})
\end{align}
where $\text{Rel}(i,j,\hat{a},\hat{b})$ is the relation between position (i,j) and position ($\hat{a}$, $\hat{b}$). With this definition, local self-attention is a special case. 

For example, SASA~\cite{ramachandran2019standalone} writes this as \begin{align}
\label{eq_sasa}
    Y_{i,j} &=  \sum_{a,b \in \mathcal{N}_{k}(i,j)} \text{Softmax}_{ab}(q_{ij}^{T}k_{ab} + q_{ij}r_{a-i, b-j})v_{ab}
\end{align}
where $q$, $k$ and $v$ are linear projections of input feature $x$, and $r_{a-i, b-j}$ is the relative positional embedding of $(i,j)$ and $(a,b)$. 

We now consider several specific works using local self-attention as basic neural network blocks

SASA~\cite{ramachandran2019standalone} suggests that using self-attention to collect global information is too computationally intensive and instead adopts local self-attention to replace all spatial convolution in a CNN. The authors show that doing so improves speed, number of parameters and quality of results. They also explores the behavior of positional embedding and show that relative positional embeddings~\cite{shaw2018relative_position} are suitable. Their work also studies how to combinie local self-attention with convolution.

LR-Net~\cite{hu2019lrnet} appeared concurrently with SASA. It also studies how to model local relationships by using local self-attention. A comprehensive study  probed the effects of positional embedding, kernel size, appearance composability and adversarial attacks.

SAN~\cite{Zhao_2020_SAN} explored two modes, pairwise and patchwise, of utilizing attention for local feature aggregation. It proposed a novel vector attention adaptive both in  content and channel, and assessed its effectiveness both theoretically and practically. In addition to providing significant improvements in the image domain, it also has been proven useful in 3D point cloud processing~\cite{zhao2020pointtransformer}.

\subsubsection{Vision Transformers}
%

\begin{figure}[t]
    \centering
    \includegraphics[width=\linewidth]{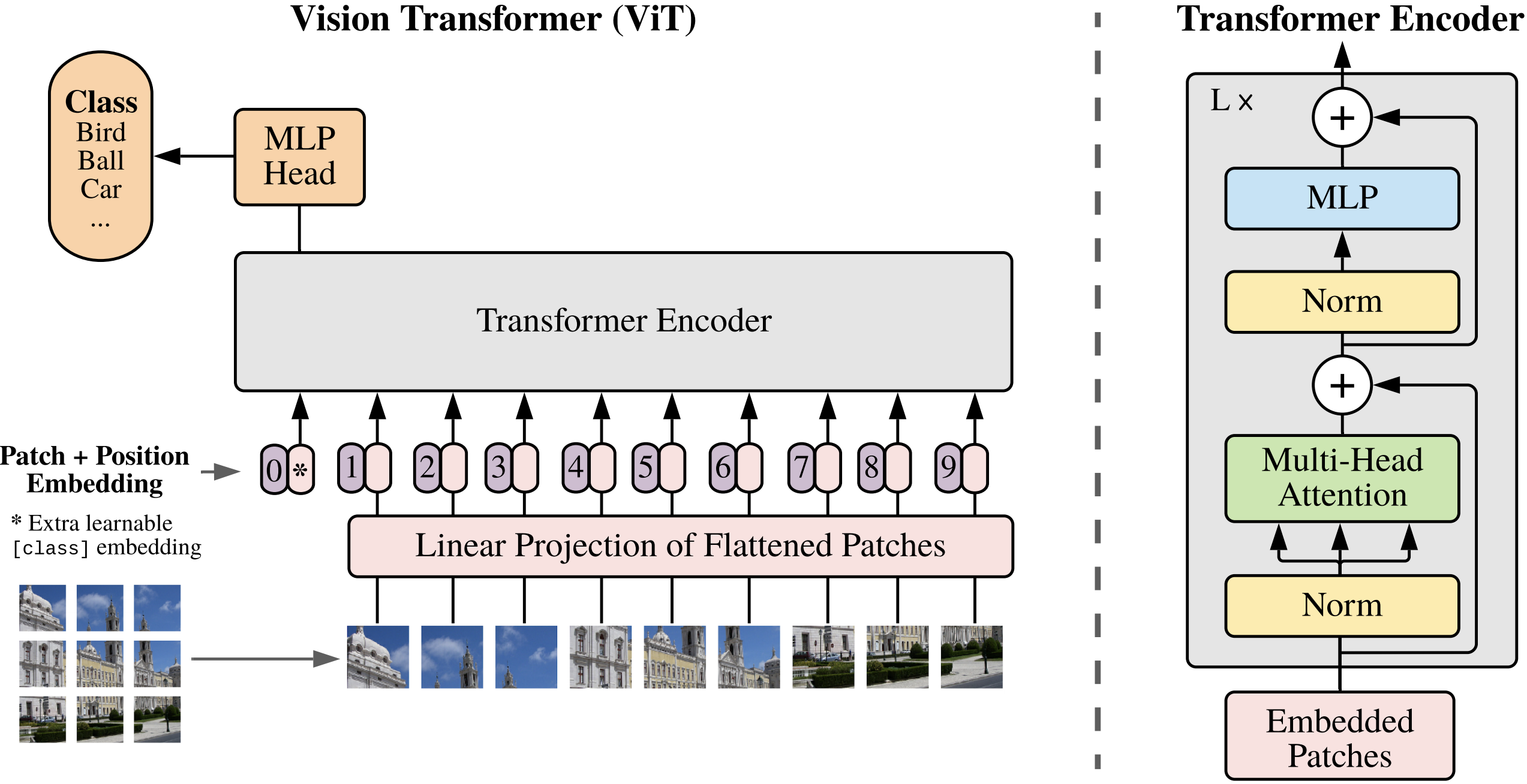}
    \caption{Vision transformer~\cite{dosovitskiy2020vit}. Left: architecture. Vision transformer first splits the image into different patches and projects them into feature space where a transformer encoder processes them to produce the final result.  Right: basic vision transformer block with   multi-head attention core. Figure is taken from~\cite{dosovitskiy2020vit}. 
    }
    \label{fig:vit}
\end{figure}

Transformers have had great success in natural language processing~\cite{bahdanau2016neural, vaswani2017attention, lin2017structured, devlin2019bert, choromanski2021rethinking, brown2020language_gpt3}. Recently, iGPT~\cite{pmlr-v119-chen20s-igpt} and DETR~\cite{carion2020endtoend_detr} demonstrated the huge potential for transformer-based models in computer vision. Motivated by this, Dosovitskiy et al~\cite{dosovitskiy2020vit} proposed the \emph{vision transformer} (ViT) which is the first pure transformer architecture for image processing. It is capable of  achieving comparable results to modern convolutional neural networks.

\begin{figure}[t]
    \centering
    \includegraphics[width=\linewidth]{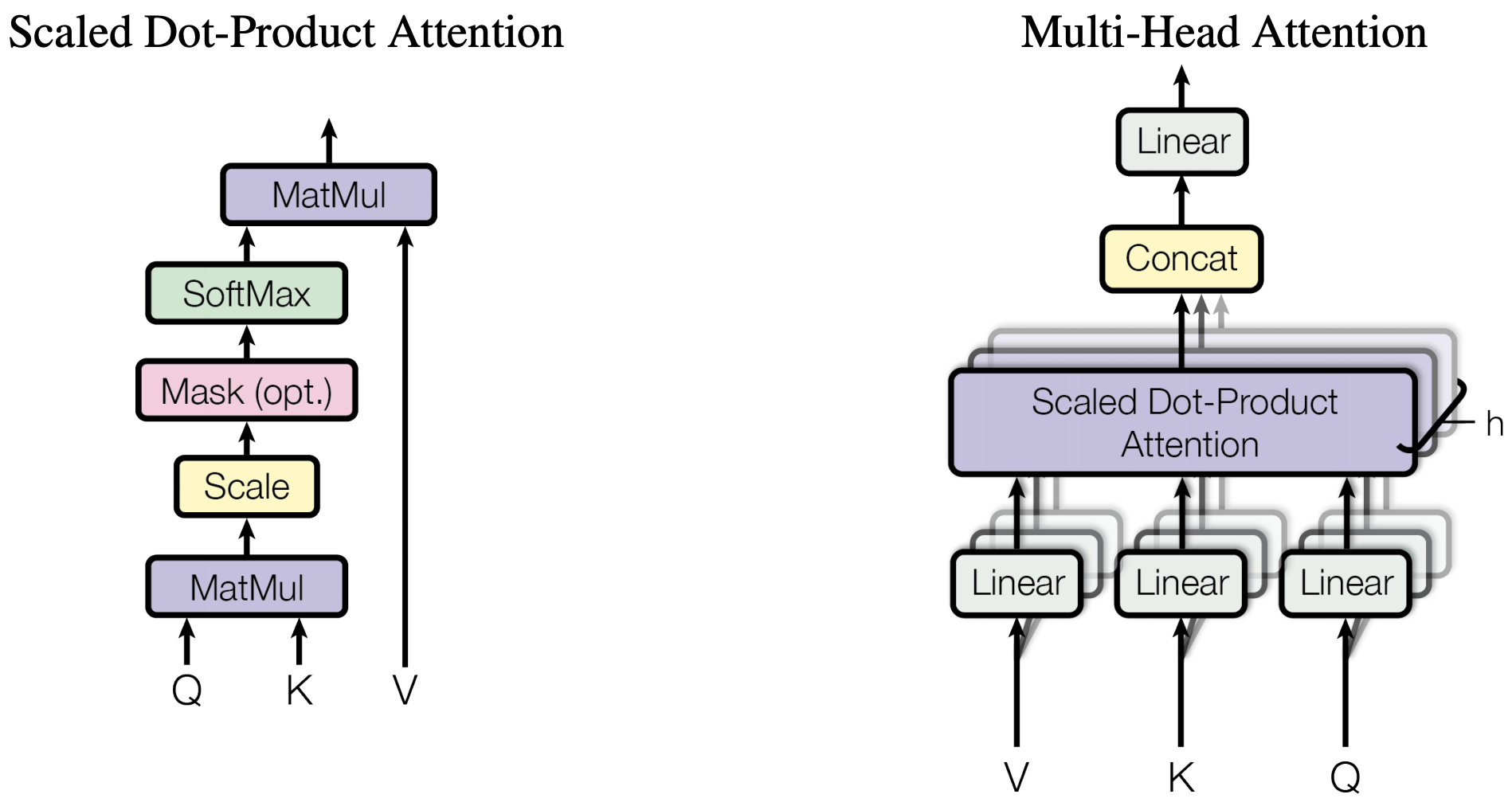}
    \caption{Left: Self-attention. Right: Multi-head self-attention. Figure from~\cite{vaswani2017attention}. 
    }
    \label{fig:multi_head_attention}
\end{figure}

As  Fig~\ref{fig:vit} shows, the main part of ViT is the multi-head attention (MHA) module. MHA takes a sequence as input. It first concatenates a class token with the input feature $F \in \mathcal{R}^{N \times C}$, where $N$ is the number of pixels.  Then it gets $Q,K \in \mathcal{R}^{N \times C'}$ and $V \in \mathcal{R}^{N \times C}$ by linear projection. Next, $Q$, $K$ and $V$ are divided into $H$ heads in the channel domain and self-attention separately applied to them. The MHA approach is shown in Fig.~\ref{fig:multi_head_attention}.  ViT stacks a number of MHA layers with fully connected layers, layer normalization~\cite{ba2016layer} and the GELU~\cite{hendrycks2020gelu} activation function.

\begin{figure*}[t]
    \centering
    \includegraphics[width=\textwidth]{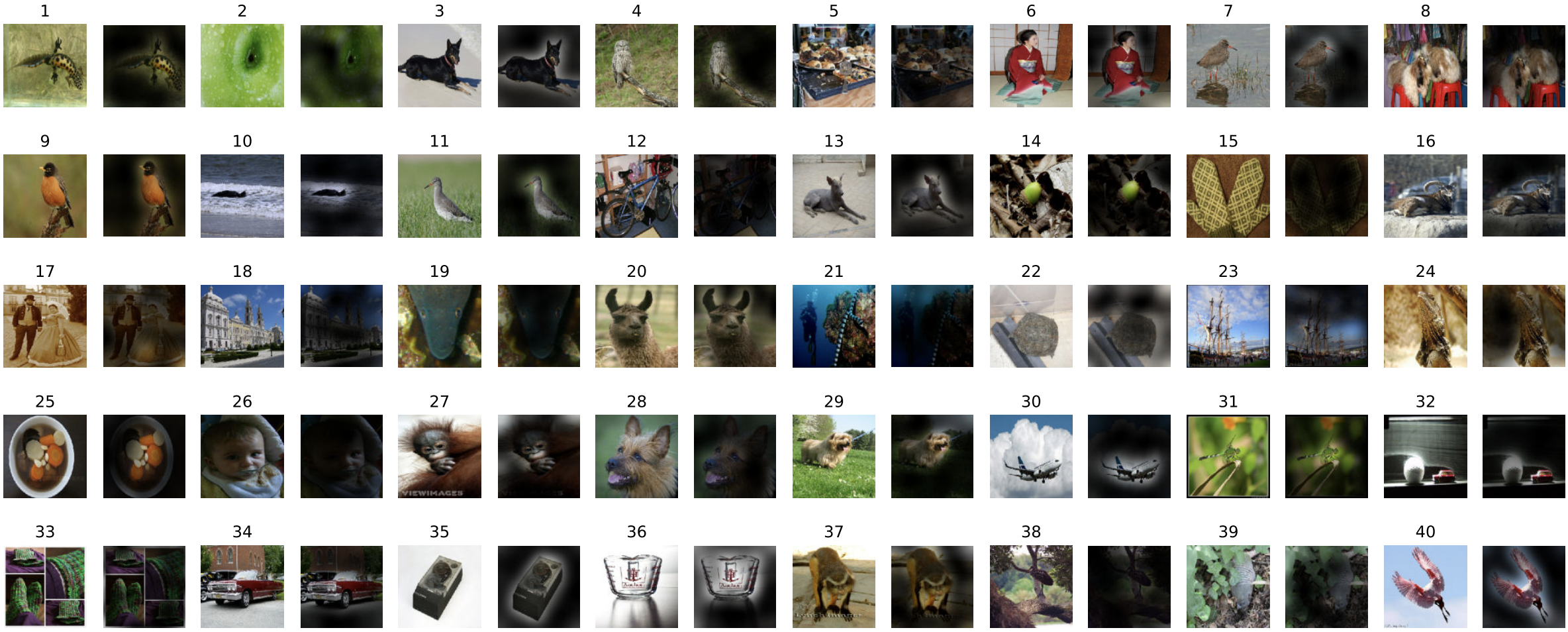}
    \caption{Attention map results from~\cite{dosovitskiy2020vit}. The network focuses on the discriminative regions of each image. Figure from~\cite{dosovitskiy2020vit}.
    }
    \label{fig:attention_map_vit}
\end{figure*}

ViT demonstrates that a pure attention-based network can achieve better results than a convolutional neural network especially for large datasets such as JFT-300~\cite{sun2017revisiting} and ImageNet-21K~\cite{Deng2009Imagenet}. 

Following ViT, many transformer-based architectures such as PCT~\cite{Guo_2021_pct}, IPT~\cite{chen2021pretrained_ipt}, T2T-ViT~\cite{yuan2021tokens}, DeepViT~\cite{zhou2021deepvit},
SETR~\cite{SETR}, PVT~\cite{wang2021pyramid}, CaiT~\cite{touvron2021going}, TNT~\cite{han2021transformer_tnt}, Swin-transformer~\cite{liu2021swin}, Query2Label~\cite{liu2021query2label}, MoCoV3~\cite{chen2021empirical_mocov3}, BEiT~\cite{bao2021beit}, SegFormer~\cite{xie2021segformer}, FuseFormer~\cite{liu2021fuseformer} 
and MAE~\cite{he2021masked}
have appeared, with excellent results for many kind of visual tasks including image classification, object detection, semantic segmentation, point cloud processing, action recognition and self-supervised learning. 

A detailed survey of vision transformers is omitted here as other recent surveys~\cite{han2021survey, khan2021transformers, guo2021attention_mlp_cnn, xu2022transformers} comprehensively review the use of transformer methods for visual tasks.

\subsubsection{GENet}
Inspired by SENet, Hu et al.~\cite{hu2019gatherexcite} designed GENet to capture long-range spatial contextual information by providing a recalibration function in the spatial domain.

GENet combines part gathering and excitation operations. In the first step, it aggregates input features over large neighborhoods and  models the relationship between different spatial locations. In the second step, it first generates an attention map of the same size as the input feature map, using interpolation. Then each position in the input feature map is scaled by multiplying by the corresponding element in the attention map. This process can be described by: 
\begin{align}
    g &= f_\text{gather}(X), \\
    s &= f_\text{excite}(g) = \sigma(\text{Interp}(g)), \\
    Y &= s  X.
\end{align}
Here, $f_\text{gather}$ can take any form which captures spatial correlations, such as global average pooling or a sequence of depth-wise convolutions; $\text{Interp}(\cdot)$ denotes interpolation. 

The gather-excite module is  lightweight and can be inserted into each residual unit like an SE block. It emphasizes important features while suppressing noise.

\subsubsection{PSANet}
Motivated by success in capturing long-range dependencies in  convolutional neural networks, Zhao et al.~\cite{Zhao_2018_psanet} presented the novel  PSANet framework to aggregate global information. It models information aggregation as an information flow and proposes a bidirectional information propagation mechanism to make information flow globally. 

PSANet formulates information aggregation as:
\begin{align}
\label{eq_psa_aggr}
z_{i} = \sum_{j \in \Omega(i)} F(x_{i}, x_{j}, \Delta_{ij})x_{j} 
\end{align}
where $\Delta_{ij}$ indicates the positional relationship between $i$ and $j$. $F(x_{i}, x_{j}, \Delta_{ij})$ is a function that takes $x_{i}$, $x_{j}$ and $\Delta_{ij}$ into consideration to controls information flow from $j$ to $i$. $\Omega_{i}$ represents the aggregation neighborhood of position $i$; if we wish to capture global information, $\Omega_{i}$ should include all spatial positions.  

Due to the complexity of calculating function $F(x_{i}, x_{j}, \Delta_{ij})$, it is decomposed into an approximation:
\begin{align}
F(x_{i}, x_{j}, \Delta_{ij}) \approx F_{\Delta_{ij}}(x_{i}) + F_{\Delta_{ij}}(x_{j})
\end{align}
whereupon Eq.~\ref{eq_psa_aggr} can be simplified to:
\begin{align}
\label{eq_psa_sim}
z_{i} = \sum_{j \in \Omega(i)} F_{\Delta_{ij}}(x_{i})x_{j} + \sum_{j \in \Omega(i)} F_{\Delta_{ij}}(x_{j})x_{j}. 
\end{align}
The first term can be viewed as  collecting information at position $i$ while the second term distributes information at position $j$. Functions $F_{\Delta_{ij}}(x_{i})$ and $F_{\Delta_{ij}}(x_{j})$ can be seen as  adaptive attention weights.  

The above process aggregates global information while emphasizing the relevant features. It can be added to the end of a convolutional neural network as an effective complement to greatly improve semantic segmentation.

\begin{table*}
	\centering
	\caption{Representative temporal attention mechanisms sorted by date. ReID = re-identification, Action = action recognition. Ranges means the ranges of attention map. S or H means soft or hard attention. $g(x)$ and $f(g(x), x)$ are the attention process described by Eq.~\ref{eq_general}. (A) aggregate information via attention map. (\uppercase\expandafter{\romannumeral1}) exploit  multi-scale short-term temporal  contextual information (\uppercase\expandafter{\romannumeral2})capture long-term temporal feature dependencies (\uppercase\expandafter{\romannumeral3}) capture local temporal contexts}
\label{Tab_temporal_attentions}
\begin{tabular}{|p{2cm}|p{2cm}|p{1.5cm}|p{1cm}|p{3cm}|p{1.4cm}|p{1.2cm}|p{0.8cm}|p{1.0cm}|}
\hline
\textbf{Category} & \textbf{Method} & \textbf{Publication} & \textbf{Tasks} & \textbf{$\bm{g(x)}$} & \textbf{$\bm{f(g(x), x)}$} & \textbf{Ranges}  & \textbf{S or H} & \textbf{Goals} \\
\hline
Self-attention based methods & GLTR~\cite{li2019global} & ICCV2019 & ReID &{ dilated 1D Convs -> self-attention in temporal dimension } &{(A)} &{(0,1)} &{S} & {(\uppercase\expandafter{\romannumeral1}), (\uppercase\expandafter{\romannumeral2}).} \\ 
\hline
Combine local attention and global attention & TAM~\cite{liu2020tam} & Arxiv2020 & Action &{ a)local: global spatial average pooling -> 1D Convs, b) global: global spatial average pooling -> MLP -> adaptive convolution } &{(A)} &{(0,1)} &{S} & {(\uppercase\expandafter{\romannumeral2}), (\uppercase\expandafter{\romannumeral3}).} \\ 
\hline
\end{tabular}
\end{table*}

\subsection{Temporal Attention}

Temporal attention can be seen as a dynamic time selection mechanism determining \emph{when to pay attention}, 
and is thus usually used for video processing.
Previous works~\cite{li2019global,liu2020tam} often emphasise
how to capture both short-term and long-term cross-frame feature dependencies. 
Here, 
we first summarize representative temporal attention mechanisms 
and specify process $g(x)$ and $f(g(x), x)$ described as Eq.~\ref{eq_general}
in Tab.~\ref{Tab_temporal_attentions}, and then discuss various such mechanisms \gmh{according to the order in Fig.~\ref{fig:total_attention_summary}}.

\subsubsection{Self-attention and variants}

RNN and temporal pooling  or weight learning 
have been widely used in work on video representation learning to capture interaction between frames, but these methods have limitations in terms of either efficiency or temporal relation modeling. 

To overcome them,
Li et al.~\cite{li2019global} proposed a \emph{global-local temporal representation} (GLTR)
to exploit multi-scale temporal cues in a video sequence.
GLTR consists of a \emph{dilated temporal pyramid} (DTP) for local temporal context learning and a \emph{temporal self attention} module for capturing global temporal interaction. DTP adopts dilated convolution with dilatation rates increasing progressively to cover various temporal ranges, and then concatenates the various outputs to aggregate multi-scale information. Given  input frame-wise features $F = \{f_1, \dots f_T\}$, DTP can be written as:
\begin{align}
    \{f_1^{(r)}, \dots, f_T^{(r)}\} &= \text{DConv}^{(r)}(F) \\
    f'_t &= [f_t^{(1)};\dots f_t^{(2^{n-1})}\dots;f_t^{(2^{N-1})}]
\end{align}
where $\text{DConv}^{(r)}(\cdot)$ denotes dilated convolution with dilation rate $r$. The self-attention mechanism  adopts convolution layers followed by batch normalization and ReLU activation to generate the query $Q \in \mathbb{R}^{d\times T}$, the key $K \in \mathbb{R}^{d\times T}$ and the value $V \in \mathbb{R}^{d\times T}$ based on the input feature map $F' = \{f'_1, \dots f'_T\}$, which can be  written as
\begin{align}
    F_\text{out} = g(V\text{Softmax}(Q^TK)) + F'
\end{align}
where $g$ denotes a linear mapping implemented by a convolution. 

The short-term temporal contextual information
from neighboring frames
helps to distinguish visually similar regions
while the long-term temporal information 
serves to overcome occlusions and noise. 
GLTR combines the advantages of both modules, 
 enhancing  representation capability
and suppressing noise. 
It can \mtj{be} incorporated into 
any state-of-the-art CNN backbone to
learn a global descriptor for a whole
video. 
However, the self-attention mechanism
has quadratic time complexity, 
 limiting its application.

\subsubsection{TAM}
To capture complex temporal relationships
 both  efficiently and  flexibly,
Liu et al.~\cite{liu2020tam} proposed a \emph{temporal adaptive module} (TAM).
It adopts an adaptive kernel instead of self-attention to capture  global contextual information, with lower time complexity 
than GLTR~\cite{li2019global}.

TAM has two branches, a local branch and a global branch. Given the input feature map $X\in \mathbb{R}^{C\times T\times H\times W}$,  global spatial average pooling $\text{GAP}$ is first applied to the feature map to ensure TAM has a low computational cost. Then the local branch in TAM employs several 1D convolutions with ReLU nonlinearity across the temporal domain to produce location-sensitive importance maps for enhancing frame-wise features.
The local branch can be written as
\begin{align}
    s &= \sigma(\text{Conv1D}(\delta(\text{Conv1D}(\text{GAP}(X)))))\\
    X^1 &= s X.
\end{align}
Unlike the local branch, the global branch is location invariant and focuses on generating a channel-wise adaptive kernel based on global temporal information in each channel. For the $c$-th channel, the  kernel can be written as
\begin{align}
    \Theta_c = \text{Softmax}(\text{FC}_2(\delta(\text{FC}_1(\text{GAP}(X)_c)))) 
\end{align}
where $\Theta_c \in \mathbb{R}^{K}$ and $K$ is the adaptive kernel size. Finally, TAM  convolves the adaptive kernel $\Theta$ with $X_\text{out}^1$:
\begin{align}
    Y = \Theta \otimes  X^1
\end{align}

With the help of the local branch and global branch,
TAM can capture the complex temporal structures in video and 
enhance per-frame features at low computational cost.
Due to its flexibility and lightweight design,
TAM can be added to any existing 2D CNNs.

\subsection{Branch Attention}

Branch attention can be seen as a dynamic branch selection mechanism: \emph{which to pay attention to},  used with a multi-branch structure. We first summarize  representative branch attention mechanisms 
and specify process $g(x)$ and $f(g(x), x)$ described as Eq.~\ref{eq_general}
in Tab.~\ref{Tab_branch_attentions}, then discuss various ones in detail.

\begin{table*}
	\centering
	\caption{Representative branch attention mechanisms sorted by date. Cls = classification, Det=Object Detection. $g(x)$ and $f(g(x), x)$ are the attention process described by Eq.~\ref{eq_general}.Ranges means the ranges of attention map. S or H means soft or hard attention. (A) element-wise product. (B) channel-wise product. (C) aggergate information via attention. (\uppercase\expandafter{\romannumeral1})overcome the problem of vanishing gradient (\uppercase\expandafter{\romannumeral2}) dynamically fuse different branches. (\uppercase\expandafter{\romannumeral3}) adaptively select a suitable receptive field (\uppercase\expandafter{\romannumeral4}) improve the performance of standard convolution (be) dynamically fuse different convolution  kernels. }
\label{Tab_branch_attentions}
\begin{tabular}{|p{2cm}|p{2cm}|p{1.5cm}|p{1cm}|p{3cm}|p{1.4cm}|p{1.2cm}|p{0.8cm}|p{1.0cm}|}
\hline
\textbf{Category} & \textbf{Method} & \textbf{Publication} & \textbf{Tasks} & \textbf{$\bm{g(x)}$} & \textbf{$\bm{f(g(x), x)}$} & \textbf{Ranges}  & \textbf{S or H} & \textbf{Goals} \\
\hline
Combine different branches & Highway Network~\cite{srivastava2015training} & ICML2015W & Cls &{linear layer -> sigmoid } & (A) & (0,1) & S & {(\uppercase\expandafter{\romannumeral1}), (\uppercase\expandafter{\romannumeral2}). } \\ 
\cline{2-9}
& SKNet~\cite{li2019selective} & CVPR2019 & Cls &{global average pooling -> MLP -> softmax} & (B) & (0,1) & S & {(\uppercase\expandafter{\romannumeral2}), (\uppercase\expandafter{\romannumeral3})} \\ 
\hline
Combine different convolution kernels & CondConv~\cite{yang2020condconv} & NeurIPS2019 & Cls, Det &{global average pooling -> linear layer -> sigmoid} & (C) & (0,1) & S & {(\uppercase\expandafter{\romannumeral4}), (\uppercase\expandafter{\romannumeral5}).} \\
\hline

\end{tabular}
\end{table*}

\subsubsection{Highway networks}

Inspired by the \emph{long short term memory} network, Srivastava et al.~\cite{srivastava2015training} proposed \emph{highway networks} that employ adaptive gating mechanisms to enable information flows across layers to address the problem of training very deep networks.

Supposing a plain neural network consists of $L$ layers, and $H_l(X)$ denotes a non-linear transformation on the $l$-th layer, a highway network can be expressed as
\begin{align}
\label{eq_highway}
    Y_l &= H_l(X_l) T_l(X_l) + X_l  (1-T_l(X_l))  \\
    T_l(X) &= \sigma(W_l^TX+b_l)
\end{align}
where $T_l(X)$ denotes the transform gate regulating the information flow for the $l$-th layer. $X_l$ and $Y_l$ are the inputs and outputs of the $l$-th layer. 

The gating mechanism and skip-connection structure make it possible to directly train very deep highway networks using simple gradient descent methods. Unlike fixed skip-connections, the gating mechanism adapts to the input, which helps to route information across layers. A highway network can be incorporated in any CNN.

\subsubsection{SKNet}


Research in the neuroscience community suggests that visual cortical neurons adaptively adjust the sizes  of their receptive fields (RFs)  according to the input stimulus~\cite{spillmann2015beyond}. This inspired Li et al.~\cite{li2019selective} to propose an automatic selection operation called \emph{selective kernel} (SK) convolution. 

SK convolution is implemented using three operations: split, fuse and select. During split, transformations with different kernel sizes are applied to the feature map to obtain different sized RFs. Information from all branches is then fused together via  element-wise summation to compute the gate vector. This is used to control information flows from the multiple branches. Finally, the output feature map is obtained by aggregating feature maps for all branches, guided by the gate vector. This can be expressed as:
\begin{align}
    U_k &= F_k(X) \qquad k=1,\dots,K\\
    U &= \sum_{k=1}^K U_k \\
    z &= \delta(\text{BN}(W \text{GAP} (U))) \\
    s_k^{(c)} &= \frac{e^{W_k^{(c)} z}}{\sum_{k=1}^K e^{W_k^{(c)} z}} \quad k=1,\dots,K,\quad c=1,\dots,C \\
    Y &= \sum_{k=1}^K s_k  U_k
\end{align}
Here, each transformation $F_k$ has a unique kernel size to provide different scales of information for each branch. For efficiency, $F_k$ is  implemented by grouped or depthwise convolutions followed by dilated convolution, batch normalization and ReLU activation in sequence. 
$t^{(c)}$ denotes the $c$-th element of vector $t$, or the $c$-th row of matrix $t$.  

SK convolutions enable the network to adaptively adjust neurons' RF sizes according to the input, giving a notable improvement in results at little computational cost. The gate mechanism in SK convolutions is used to fuse information from multiple branches. Due to its lightweight design, SK convolution can be applied to any CNN backbone by replacing all  large kernel convolutions.   ResNeSt~\cite{zhang2020resnest} also adopts this attention mechanism to improve the CNN backbone in a more general way, giving excellent results on ResNet~\cite{resnet} and ResNeXt~\cite{resnext}.

\subsubsection{CondConv}

A basic assumption in CNNs is that all convolution kernels are the same. Given this, the typical way to enhance the representational power of a network is to increase its depth or width, which introduces significant extra computational cost. In order to more efficiently increase the capacity of convolutional neural networks, Yang et al.~\cite{yang2020condconv} proposed a novel multi-branch operator called CondConv. 

\begin{figure}[t]
    \centering
    \includegraphics[width=\linewidth]{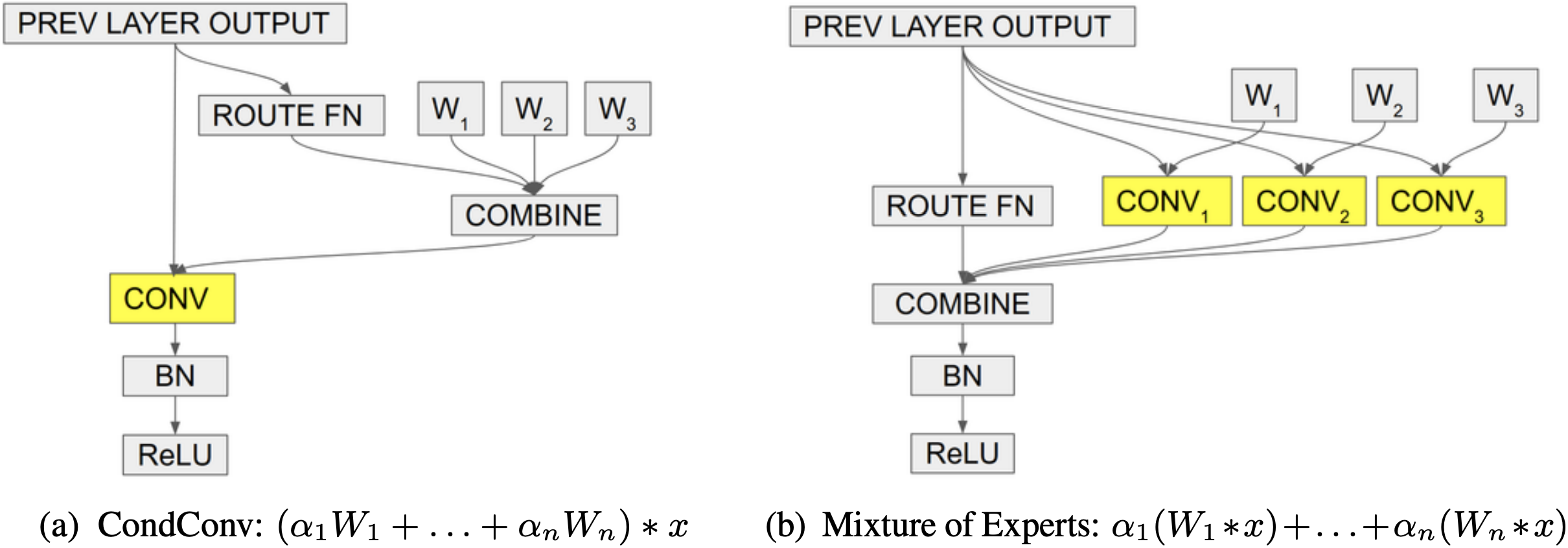}
    \caption{
    CondConv~\cite{yang2020condconv}. (a) CondConv first combines different convolution kernels and then uses the combined kernel for convolution. (b) Mixture of experts first uses multiple convolution kernels for convolution and then merges the results. While (a) and (b) are equivalent, (a) has much lower computational cost. Figure  is taken from~\cite{yang2020condconv}. 
    }
    \label{fig:CondConv}
\end{figure}

An ordinary convolution can be written
\begin{align}
    Y = W * X
\end{align}
where $*$ denotes convolution. The learnable parameter $W$ is the same for all samples.  CondConv adaptively combines multiple convolution kernels and can be written as:
\begin{align}
    Y = (\alpha_{1} W_{1} + \dots + \alpha_{n} W_{n}) * X 
\end{align}
Here, $\alpha$ is a learnable weight vector computed by 
\begin{align}
\label{eq.condconv_weight}
    \alpha = \sigma(W_{r}(\text{GAP}(X))) 
\end{align}
This process is equivalent to an ensemble of multiple experts, as shown in Fig.~\ref{fig:CondConv}. 

CondConv makes full use of the advantages of the multi-branch structure using a branch attention method with little computing cost. It presents a novel manner to efficiently increase the capability of networks.

\subsubsection{Dynamic Convolution}

The extremely low computational cost of lightweight CNNs constrains the depth and width of the networks, further decreasing their representational power. To address the above problem, Chen et al.~\cite{chen2020dynamic} proposed \emph{dynamic convolution}, a novel operator design that increases  representational power with  negligible additional computational cost and does not change the width or depth of the network in parallel with CondConv~\cite{yang2020condconv}. 

Dynamic convolution uses $K$ parallel convolution kernels of the same  size and input/output dimensions instead of one kernel per layer. Like SE blocks, it adopts a squeeze-and-excitation mechanism to generate the attention weights for the different convolution kernels. These kernels are then aggregated dynamically by weighted summation and applied to the input feature map $X$:
\begin{align}
    s & = \text{softmax} (W_{2} \delta (W_{1}\text{GAP}(X))) \\
    \text{DyConv} &= \sum_{i=1}^{K} s_k \text{Conv}_k \\
    Y &= \text{DyConv}(X)
\end{align}
Here the convolutions are combined by summation of weights and biases of convolutional kernels. 

Compared to applying convolution to the feature map, the computational cost of squeeze-and-excitation and weighted summation is extremely low. Dynamic convolution thus provides an efficient operation to improve  representational power and can be easily used as a replacement for any convolution.  

\subsection{Channel \& Spatial Attention}

Channel \& spatial attention combines the advantages of 
channel attention and spatial attention. 
It adaptively selects both important objects and regions~\cite{ChenZXNSLC17}. 
The \emph{residual attention network}~\cite{wang2017residual}  pioneered the field of channel \& spatial attention,
 emphasizing the importance of informative features in both spatial and channel dimensions. 
It adopts a bottom-up structure consisting of several convolutions to produce a 3D (height, width, channel) attention map. 
However, it has  high computational cost
and limited receptive fields.

To leverage global spatial information 
later works~\cite{woo2018cbam,park2018bam} enhance
discrimination of features by introducing global average pooling,
as well as decoupling channel attention 
and spatial channel attention for computational efficiency.
Other works~\cite{Fu_2019_danet,zhang2020relation} apply
self-attention mechanisms for channel \& spatial attention
to explore pairwise interaction.
Yet further works~\cite{Liu_2020_scnet,hou2020strip} 
adopt the
spatial-channel attention mechanism to enlarge the receptive field.

Representative channel \& spatial attention mechanisms and specific process $g(x)$ and $f(g(x), x)$ described as Eq.~\ref{eq_general}
 are in given Tab.~\ref{Tab_channel_spatial_attentions}; we next discuss various ones in detail.

\begin{table*}
	\centering
	\caption{Representative channel \& spatial attention mechanisms sorted by date. Cls = classification,  ICap = image captioning, Det = detection, Seg = segmentation, ISeg = instance segmentation, KP = keypoint detection, ReID = re-identification.$g(x)$ and $f(g(x), x)$ are the attention process described by Eq.~\ref{eq_general}. Ranges means the ranges of attention map. S or H means soft or hard attention. (A) element-wise product. (B) aggregate information via attention map.(\uppercase\expandafter{\romannumeral1}) focus the network  on the  discriminative region, (\uppercase\expandafter{\romannumeral2}) emphasize important channels, (\uppercase\expandafter{\romannumeral3}) capture long-range information, (\uppercase\expandafter{\romannumeral4}) capture cross-domain interaction  between any two domains. }
\label{Tab_channel_spatial_attentions}
\begin{tabular}{|p{2cm}|p{2cm}|p{1.5cm}|p{1cm}|p{3cm}|p{1.4cm}|p{1.2cm}|p{0.8cm}|p{1.0cm}|}
\hline
\textbf{Category} & \textbf{Method} & \textbf{Publication} & \textbf{Tasks} & \textbf{$\bm{g(x)}$} & \textbf{$\bm{f(g(x), x)}$} & \textbf{Ranges}  & \textbf{S or H} & \textbf{Goals} \\
\hline
Jointly predict channel \& spatial attention map & Residual Attention \cite{wang2017residual} & CVPR2017 & Cls &{top-down network -> bottom down network -> $1\times 1$ Convs -> Sigmoid} &{(A)} & {(0,1)} &{S} & {(\uppercase\expandafter{\romannumeral1}), (\uppercase\expandafter{\romannumeral2})} \\
\cline{2-9}
& SCNet~\cite{Liu_2020_scnet} & CVPR2020 & Cls, Det, ISeg, KP &{top-down network -> bottom down network -> identity add -> sigmoid} &{(A)} & {(0,1)} &{S}& {(\uppercase\expandafter{\romannumeral2}), (\uppercase\expandafter{\romannumeral3})} \\ 
\cline{2-9}
& Strip Pooling~\cite{hou2020strip} & CVPR2020 & Seg &{a)horizontal/vertical global pooling -> 1D Conv -> point-wise summation -> $1\times 1$ Conv -> Sigmoid} &{(A)} & {(0,1)} &{S}& {(\uppercase\expandafter{\romannumeral1}), (\uppercase\expandafter{\romannumeral2}), (\uppercase\expandafter{\romannumeral3})} \\ 
\hline
Separately predict channel \& spatial attention maps & SCA-CNN~\cite{ChenZXNSLC17} & CVPR2017 & ICap &{a)spatial: fuse hidden state -> $1\times 1$ Conv -> Softmax, b)channel: global average pooling -> MLP -> Softmax } &{(A)} & {(0,1)} &{S} & {(\uppercase\expandafter{\romannumeral1}), (\uppercase\expandafter{\romannumeral2}), (\uppercase\expandafter{\romannumeral3})  } \\ 
\cline{2-9}
 & CBAM~\cite{woo2018cbam} & ECCV2018 & Cls, Det &{a)spatial: global pooling in channel dimension-> Conv -> Sigmoid, b)channel: global pooling in spatial dimension -> MLP -> Sigmoid} &{(A)} & {(0,1)} &{S} & {(\uppercase\expandafter{\romannumeral1}), (\uppercase\expandafter{\romannumeral2}), (\uppercase\expandafter{\romannumeral3})} \\ 

\cline{2-9}

& BAM~\cite{woo2018cbam} & BMVC2018 & Cls, Det &{a)spatial: dilated Convs, b)channel: global average pooling -> MLP, c)fuse  two branches} &{(A)} & {(0,1)} &{S} & {(\uppercase\expandafter{\romannumeral1}), (\uppercase\expandafter{\romannumeral2}), (\uppercase\expandafter{\romannumeral3})} \\ 
\cline{2-9}
& scSE~\cite{roy2018recalibrating} & TMI2018 & Seg &{a)spatial: $1\times 1$ Conv -> Sigmoid,  b)channel: global average pooling -> MLP -> Sigmoid, c)fuse two branches } &{(A)} & {(0,1)} &{S}& {(\uppercase\expandafter{\romannumeral1}), (\uppercase\expandafter{\romannumeral2}), (\uppercase\expandafter{\romannumeral3})} \\
\cline{2-9}
& Dual Attention~\cite{Fu_2019_danet} & CVPR2019 & Seg &{a)spatial: self-attention in spatial dimension, b)channel: self-attention in channel dimension, c) fuse two branches} &{(B)} & {(0,1)} &{S}& {(\uppercase\expandafter{\romannumeral1}), (\uppercase\expandafter{\romannumeral2}), (\uppercase\expandafter{\romannumeral3})} \\ 
\cline{2-9}
& RGA~\cite{zhang2020relation} & CVPR2020 & ReID &{use self-attention to capture pairwise relations -> compute attention maps with the input and  relation vectors} &{(A)} & {(0,1)} &{S}& { (\uppercase\expandafter{\romannumeral1}), (\uppercase\expandafter{\romannumeral2}), (\uppercase\expandafter{\romannumeral3})} \\
\cline{2-9}
& Triplet Attention~\cite{misra2021rotate} & WACV2021 & Cls, Det &{compute  attention maps for pairs of domains -> fuse different branches} &{(A)} & {(0,1)} &{S}& {(\uppercase\expandafter{\romannumeral1}), (\uppercase\expandafter{\romannumeral4}) } \\ 

\hline
\end{tabular}
\end{table*}

\subsubsection{Residual Attention Network}


Inspired by the success of ResNet~\cite{resnet}, 
Wang et al.~\cite{wang2017residual} proposed
the very deep convolutional \emph{residual attention network} (RAN) by 
combining an attention mechanism with residual connections. 

Each attention module stacked in a residual attention network 
can be divided into a mask branch and a trunk branch. 
The trunk branch processes features,
and can be implemented by any state-of-the-art structure
including a pre-activation residual unit and an inception block.
The mask branch uses a bottom-up top-down structure
to learn a mask of the same size that 
softly weights output features from the trunk branch. 
A sigmoid layer normalizes the output to $[0,1]$ after two $1\times 1$ convolution layers. Overall the residual attention mechanism can be written as
\begin{align}
s &= \sigma(\text{Conv}_{2}^{1\times 1}(\text{Conv}_{1}^{1\times 1}( h_\text{up}(h_\text{down}(X))))) \\
X_{out} &= s f(X) + f(X)
\end{align}
where $h_\text{up}$ is a bottom-up structure, 
using max-pooling several times after residual units
to increase the receptive field, while
$h_\text{down}$ is the top-down part using 
linear interpolation to keep the output size the 
same as the input feature map. 
There are also skip-connections between the two parts,
which are omitted from the formulation.
$f$ represents the trunk branch
which can be any state-of-the-art structure.

Inside each attention module, a
bottom-up top-down feedforward structure models
both spatial and cross-channel dependencies, 
 leading to a consistent performance improvement. 
Residual attention can be incorporated into
any deep network structure in an end-to-end training fashion.
However, the proposed bottom-up top-down structure fails to leverage global spatial information.  
Furthermore, directly predicting a 3D attention map  has high computational cost.

\subsubsection{CBAM}

To enhance informative channels as well as important regions, 
Woo et al.~\cite{woo2018cbam} proposed the
\emph{convolutional block attention module} (CBAM)
which stacks channel attention and spatial attention in series. 
It decouples the channel attention map and spatial attention map for computational efficiency,
and leverages spatial global information by introducing 
global pooling.

CBAM has two sequential sub-modules, channel and spatial. 
Given an input feature map $X\in \mathbb{R}^{C\times H \times W}$
it sequentially infers 
a 1D channel attention vector $s_c\in \mathbb{R}^C$ and 
a 2D spatial attention map $s_s\in \mathbb{R}^{H\times W}$.
The formulation of the channel attention sub-module is similar to that of an SE block,
except that it adopts more than one type of pooling operation to aggregate global information. 
In detail, it has two parallel branches using max-pool and avg-pool operations:
\begin{align}
F_\text{avg}^c &= \text{GAP}^s(X) \\
F_\text{max}^c &= \text{GMP}^s(X) \\
s_c &= \sigma(W_2\delta(W_1 F_\text{avg}^c)+W_2\delta(W_1 F_\text{max}^c))\\
M_c(X) &= s_c X
\end{align}
where $\text{GAP}^s$ and $\text{GMP}^s$ denote global average pooling and  global max pooling operations in the spatial domain. 
The spatial attention sub-module models the spatial relationships of features, and is complementary to channel attention. Unlike channel attention, it applies a convolution layer with a large kernel to generate the attention map
\begin{align}
F_\text{avg}^s &= \text{GAP}^c(X) \\
F_\text{max}^s &= \text{GMP}^c(X) \\
s_s &= \sigma(\text{Conv}([F_\text{avg}^s;F_\text{max}^s]))\\
M_s(X) &= s_s X
\end{align}
where $\text{Conv}(\cdot)$ represents a convolution operation, while $\text{GAP}^c$ and $\text{GMP}^c$ are global pooling operations in the channel domain. $[]$ denotes  concatenation over channels. The overall attention process can be summarized as 
\begin{align}
    X' &= M_c(X) \\
    Y &= M_s(X')
\end{align}

Combining channel attention and spatial attention sequentially,
CBAM can utilize both spatial and cross-channel relationships of features to tell the network \emph{what} to focus on and \emph{where} to focus.
To be more specific, it emphasizes useful channels as well as enhancing informative local regions. 
Due to its lightweight design, CBAM can be integrated into any CNN architecture seamlessly with negligible additional cost. 
Nevertheless, there is still room for improvement in the channel \& spatial attention mechanism. For instance, CBAM adopts a convolution to produce the spatial attention map, so the spatial sub-module may suffer from a limited receptive field.

\subsubsection{BAM}

At the same time as CBAM, Park et al.~\cite{park2018bam} proposed the \emph{bottleneck attention module} (BAM), aiming
to efficiently improve the representational capability of networks. 
It uses dilated convolution to enlarge the receptive field of the spatial attention sub-module, and build a \emph{bottleneck structure} as suggested  by ResNet to save computational cost.

For a given input feature map $X$, BAM infers the channel attention $s_c \in \mathbb{R}^C$ and spatial attention $s_s\in \mathbb{R}^{H\times W}$ in two parallel streams, then sums the two attention maps after resizing both branch outputs to $\mathbb{R}^{C\times H \times W}$. The channel attention branch, like an SE block, applies global average pooling to the feature map to aggregate global information, and then uses an MLP with channel dimensionality reduction. In order to utilize contextual information effectively, the spatial attention branch combines a bottleneck structure and dilated convolutions. Overall, BAM can be written as
\begin{align}
    s_c &= \text{BN}(W_2(W_1\text{GAP}(X)+b_1)+b_2) \\
    s_s &= \text{BN}(\text{Conv}_{2}^{1\times 1}(\text{DC}_{2}^{3\times 3}(\text{DC}_{1}^{3\times 3}(\text{Conv}_{1}^{1\times 1}(X))))) \\
    s &= \sigma(\text{Expand}(s_s)+\text{Expand}(s_c)) \\
    Y &= s X+X
\end{align}
where $W_i$, $b_i$ denote  weights and biases of fully connected layers respectively, $\text{Conv}_{1}^{1\times 1}$ and $\text{Conv}_{2}^{1\times 1}$ are convolution layers  used for channel reduction. $\text{DC}_i^{3\times 3}$ denotes a dilated convolution with $3\times 3$ kernel,  applied to utilize contextual information effectively. $\text{Expand}$ expands the attention maps $s_s$ and $s_c$ to $\mathbb{R}^{C\times H\times W}$.

BAM can emphasize or suppress features in both spatial and channel dimensions, as well as improving the representational power. Dimensional reduction applied to both channel and spatial attention branches enables it to be integrated with any convolutional neural network with little extra computational cost. However, although dilated convolutions enlarge the receptive field effectively, it still fails to capture long-range contextual information as well as encoding cross-domain relationships.

\subsubsection{scSE}

To aggregate global spatial information,
an SE block applies global pooling to the feature map.
However, it ignores pixel-wise spatial information,
which is important in \gmh{dense prediction tasks}.
Therefore, Roy et al.~\cite{roy2018recalibrating} proposed
\emph{spatial and channel SE blocks} (scSE). 
Like BAM, spatial SE blocks are used, complementing SE blocks, 
to provide spatial attention weights to focus on important regions.

Given the input feature map $X$, two parallel modules, spatial SE and channel SE, are applied to feature maps to encode spatial and channel information respectively. The channel SE module is an ordinary SE block, while the spatial SE module adopts $1\times 1$ convolution for spatial squeezing. The outputs from the two modules are fused. The overall process can be written as
\begin{align}
    s_c & = \sigma (W_{2} \delta (W_{1}\text{GAP}(X))) \\
    X_\text{chn} & = s_c  X \\
    s_s &= \sigma(\text{Conv}^{1\times 1}(X)) \\
    X_\text{spa} & = s_s  X \\
    Y &= f(X_\text{spa},X_\text{chn})  
\end{align}
where $f$ denotes the fusion function, which can be  maximum, addition, multiplication or concatenation. 

The proposed scSE block combines channel and spatial attention to
enhance features as well as 
capturing pixel-wise spatial information.
Segmentation tasks are greatly benefited as a result.
The integration of an scSE block in F-CNNs makes a consistent improvement in semantic segmentation at negligible extra cost. 

\subsubsection{Triplet Attention}

In CBAM and BAM,  channel attention and spatial attention are computed independently,  ignoring  relationships between these two domains~\cite{misra2021rotate}.
Motivated by spatial attention, Misra et al.~\cite{misra2021rotate} proposed \emph{triplet attention},
a lightweight but effective attention mechanism
 to capture cross-domain interaction. 

Given an input feature map $X$, triplet attention uses three branches, each of which plays a role in capturing cross-domain interaction between any two domains from $H$, $W$ and $C$. In each branch,  rotation operations along different axes are applied to the input first, and then a Z-pool layer is responsible for aggregating information in the zeroth dimension. Finally, a standard convolution layer with kernel size $k\times k$  models the relationship between the last two domains. This process can be written as 
\begin{align}
X_1 &= \text{Pm}_1(X) \\
X_2 &= \text{Pm}_2(X) \\
s_0 &= \sigma(\text{Conv}_0(\text{Z-Pool}(X))) \\
s_1 &= \sigma(\text{Conv}_1(\text{Z-Pool}(X_1))) \\
s_2 &= \sigma(\text{Conv}_2(\text{Z-Pool}(X_2))) \\
Y &= \frac{1}{3}(s_0X + \text{Pm}_1^{-1} (s_1X_1)+\text{Pm}_2^{-1}(s_2X_2 ))
\end{align}
where $\text{Pm}_1$ and $\text{Pm}_2$ denote  rotation through 90$^\circ$ anti-clockwise about the $H$ and $W$ axes respectively, while $\text{Pm}^{-1}_i$ denotes the inverse. $\text{Z-Pool}$ concatenates max-pooling and average pooling along the zeroth dimension.
\begin{align}
    Y = \text{Z-Pool}(X) &= [ \text{GMP}(X); \text{GAP}(X)]
\end{align}

Unlike CBAM and BAM, triplet attention stresses the
importance of capturing cross-domain interactions 
instead of computing spatial attention and channel attention independently. 
This helps to capture rich discriminative feature representations.
Due to its simple but efficient structure, triplet attention can be 
easily added to classical backbone networks.

\subsubsection{SimAM}

Yang et al.~\cite{pmlr-v139-simam} also stress the importance of learning attention weights that vary across both channel and spatial domains in proposing SimAM, a simple, parameter-free attention module capable of directly estimating 3D weights instead of expanding 1D or 2D weights. The design of SimAM is based on well-known neuroscience theory, thus avoiding need for manual fine tuning of the network structure. 


Motivated by the spatial suppression phenomenon~\cite{Webb_spatial_supp}, they propose that a neuron which shows suppression effects should be emphasized and define an energy function for each neuron as: 
\begin{align}
\label{eq_sim_energy}
    e_{t}(w_{t}, b_{t}, y, x_{i}) = (y_{t} - \hat{t})^{2} + \frac{1}{M-1} \sum_{i=1}^{M-1}(y_{o} - \hat{x}_{i})
\end{align} 
where $\hat{t}=w_{t}t+b_{t}$, $\hat{x}_{i} = w_{t}x_{i}+b_{t}$, and $t$ and $x_{i}$ are the target unit and all other units in the same channel; $i \in {1, \dots, N}$, and $ N= H \times W$. 

An optimal closed-form solution for Eq.~\ref{eq_sim_energy} exists:\begin{align}
\label{eq_sim_solution}
    e_{t}^{\ast} = \frac{4(\hat{\sigma}^{2} + \lambda)}{(t - \hat{\mu})^2 + 2\hat{\sigma}^{2} + 2\lambda} 
\end{align} 
where $\hat{\mu}$ is the mean of the input feature and $\hat{\sigma}^{2}$ is its variance. A sigmoid function is used to control the output range of the attention vector; an element-product is applied to get the final output:

\begin{align}
    Y = \text{Sigmoid}\left(\frac{1}{E}\right) X
\end{align} 

This \mtj{work} 
simplifies the process of designing attention and successfully proposes a novel 3-D weight parameter-free attention module based on mathematics and neuroscience theories.

\subsubsection{Coordinate attention}


An SE block aggregates global spatial information using global pooling before modeling cross-channel relationships, but neglects the importance of positional information. 
BAM and CBAM adopt convolutions to capture local relations, but fail to model long-range dependencies. 
To solve these problems,
Hou et al.~\cite{hou2021coordinate} proposed \emph{coordinate attention},
a novel attention mechanism which
embeds positional information into channel attention,
so that the network can focus on large important regions 
at little computational cost.

The coordinate attention mechanism has two consecutive steps, coordinate information embedding and coordinate attention generation. First, two spatial extents of pooling kernels encode each channel horizontally  and  vertically. In the second step, a shared $1\times 1$ convolutional transformation function is applied to the concatenated outputs of the two pooling layers. Then coordinate attention splits the resulting tensor into two separate tensors to yield attention vectors with the same number of channels for horizontal and vertical coordinates of the  input $X$ along. This can be written as 
\begin{align}
    z^h &= \text{GAP}^h(X) \\
    z^w &= \text{GAP}^w(X) \\
    f &= \delta(\text{BN}(\text{Conv}_1^{1\times 1}([z^h;z^w]))) \\
    f^h, f^w &= \text{Split}(f) \\
    s^h &= \sigma(\text{Conv}_h^{1\times 1}(f^h)) \\
    s^w &= \sigma(\text{Conv}_w^{1\times 1}(f^w)) \\
    Y &= X s^h  s^w
\end{align}
where $\text{GAP}^h$ and $\text{GAP}^w$ denote pooling functions for vertical and horizontal coordinates, and $s^h \in \mathbb{R}^{C\times 1\times W}$ and $s^w \in \mathbb{R}^{C\times H\times 1}$ represent corresponding attention weights. 

Using coordinate attention, the network can accurately obtain the position of a targeted object.
This approach has a larger receptive field than BAM and CBAM.
Like an SE block, it also models cross-channel relationships, effectively enhancing the expressive power of the learned features.
Due to its lightweight design and flexibility, 
it can be easily used in classical building blocks of mobile networks.

\subsubsection{DANet} 

In the field of scene segmentation,
encoder-decoder structures cannot make use of the global relationships 
between objects, whereas RNN-based structures 
heavily rely on the output of the long-term memorization.
To address the above problems, 
Fu et al.~\cite{Fu_2019_danet} proposed a novel framework, 
 the \emph{dual attention network} (DANet), 
for natural scene image segmentation. 
Unlike CBAM and BAM, it adopts a self-attention mechanism 
instead of simply stacking convolutions to compute the spatial attention map,
which enables the network to capture global information directly. 

DANet uses in parallel a position attention module and a channel attention module to capture feature dependencies in spatial and channel domains. Given the input feature map $X$, convolution layers are applied first in the position attention module to obtain new feature maps. Then the position attention module selectively aggregates the features at each position using a weighted sum of features at all positions, where the weights are determined by feature similarity between corresponding pairs of positions. The channel attention module has a similar form except for dimensional reduction to model cross-channel relations. Finally the outputs from the two branches are fused to obtain final feature representations. For simplicity, we reshape the feature map $X$ to $C\times (H \times W)$ whereupon the overall process can be written as 
\begin{align}
    Q,\quad K,\quad V &= W_qX,\quad W_kX,\quad W_vX \\
    Y^\text{pos} &=  X+ V\text{Softmax}(Q^TK) \\
    Y^\text{chn} &=  X+ \text{Softmax}(XX^T)X  \\
    Y &= Y^\text{pos} + Y^\text{chn}
\end{align}
where $W_q$, $W_k$, $W_v \in \mathbb{R}^{C\times C}$ are used to generate new feature maps.   

The position attention module enables
DANet to capture long-range contextual information
and adaptively integrate similar features at any scale
from a global viewpoint,
while the channel attention module is responsible for 
enhancing useful channels 
as well as suppressing noise. 
Taking spatial and channel 
relationships into consideration explicitly
improves the feature representation for scene segmentation.
However, it is computationally costly, especially for large input feature maps.

\subsubsection{RGA}

Unlike coordinate attention and DANet, which emphasise capturing long-range context, in \emph{relation-aware global attention} (RGA)~\cite{zhang2020relation}, Zhang et al. stress the importance of global structural information provided by pairwise relations, and uses it to produce attention maps. 

RGA comes in two forms,  \emph{spatial RGA} (RGA-S) and \emph{channel RGA} (RGA-C). RGA-S first reshapes the input feature map $X$ to $C\times (H\times W)$ and the pairwise relation matrix $R \in \mathbb{R}^{(H\times W)\times (H\times W)}$ is computed using 
\begin{align}
    Q &= \delta(W^QX) \\
    K &= \delta(W^KX) \\
    R &= Q^TK
\end{align}
The relation vector $r_i$ at position $i$ is defined by stacking  pairwise relations at all positions:
\begin{align}
    r_i = [R(i, :); R(:,i)]    
\end{align}
and the spatial relation-aware feature $y_i$ can be written as
\begin{align}
    Y_i = [g^c_\text{avg}(\delta(W^\varphi x_i)); \delta(W^\phi r_i)]
\end{align}
where $g^c_\text{avg}$ denotes global average pooling in the channel domain. Finally, the spatial attention score at position $i$ is given by 
\begin{align}
    a_i = \sigma(W_2\delta(W_1y_i))
\end{align}
RGA-C has the same form as RGA-S, except for taking the input feature map as a set of $H\times W$-dimensional features.

RGA uses global relations to generate the attention score for each feature node,  so provides valuable structural information and significantly enhances the representational power. RGA-S and RGA-C are flexible enough to be used in any CNN network; Zhang et al. propose  using them jointly in sequence to better capture both spatial and cross-channel relationships. 

\subsubsection{Self-Calibrated Convolutions}

Motivated by the success of group convolution, Liu et at~\cite{Liu_2020_scnet} presented \emph{self-calibrated convolution} as a means to enlarge the receptive field at each spatial location. 

Self-calibrated convolution is used together with a standard convolution. It first divides the input feature $X$ into $X_{1}$ and $X_{2}$ in the channel domain. The self-calibrated convolution first uses average pooling to reduce the input size and enlarge the receptive field:
\begin{align}
T_{1} = \text{AvgPool}_{r}(X_{1}) 
\end{align}
where $r$ is the filter size and stride. Then a convolution is used to model the channel relationship and a bilinear interpolation operator $Up$ is used to upsample the feature map: 
\begin{align}
X'_{1} = \text{Up}(\text{Conv}_2(T_{1}))
\end{align}
Next, element-wise multiplication finishes the self-calibrated process:
\begin{align}
Y'_{1} = \text{Conv}_3(X_{1}) \sigma(X_{1} + X'_{1})
\end{align}
Finally, the output feature map of is formed:
\begin{align}
Y_{1} &= \text{Conv}_4(Y'_{1}) \\
Y_2 &= \text{Conv}_1(X_2) \\
Y &= [Y_1; Y_2]
\end{align}
Such self-calibrated convolution can enlarge the receptive field of a network and improve its adaptability. It achieves excellent results in image classification and certain downstream tasks such as instance segmentation, object detection and keypoint detection.

\subsubsection{SPNet}

Spatial pooling usually operates on a small region which limits its capability to capture long-range dependencies and focus on distant regions. To overcome this, Hou et al.~\cite{hou2020strip} proposed  \emph{strip pooling}, a novel pooling method capable of encoding long-range context in either horizontal or vertical spatial domains.  

Strip pooling has two branches for horizontal  and vertical strip pooling. The horizontal strip pooling part first pools the input feature $F \in \mathcal{R}^{C \times H \times W}$ in the horizontal direction:
\begin{align}
y^1 = \text{GAP}^w (X) 
\end{align}
Then a 1D convolution with kernel size 3 is applied in $y$ to capture the relationship between different rows and channels. This is repeated $W$ times to make  the output $y_v$  consistent with the input shape:
\begin{align}
    y_h = \text{Expand}(\text{Conv1D}(y^1))
\end{align}
Vertical strip pooling is performed in a similar way. Finally, the outputs of the two branches are fused using element-wise summation to produce the attention map:
\begin{align}
s &= \sigma(\text{Conv}^{1\times 1}(y_{v} + y_{h})) \\ 
Y &= s  X
\end{align}

The strip pooling module (SPM) is further developed in the mixed pooling module (MPM). Both consider  spatial  and channel relationships to overcome the locality of convolutional neural networks.  SPNet achieves  state-of-the-art results for several complex semantic segmentation benchmarks.

\subsubsection{SCA-CNN}

As CNN features are naturally spatial, channel-wise and multi-layer, 
Chen et al.~\cite{ChenZXNSLC17} proposed a novel \emph{spatial and channel-wise attention-based convolutional neural network} (SCA-CNN). 
It was designed for the task of image captioning, and uses an encoder-decoder framework where a CNN first encodes an input image into a vector and then an LSTM decodes the vector into a sequence of words. Given an input feature map $X$ and the previous time step LSTM hidden state $h_{t-1} \in \mathbb{R}^d$, a spatial attention mechanism pays more attention to the semantically useful regions, guided by LSTM hidden state $h_{t-1}$. The  spatial attention model is:
\begin{align}
a(h_{t-1}, X) &= \tanh(\text{Conv}^{1\times1}_1(X)\oplus W_1h_{t-1}) \\
\Phi_s(h_{t-1}, X) &= \text{Softmax}(\text{Conv}^{1\times1}_2(a(h_{t-1}, X)))    
\end{align}
where $\oplus$ represents  addition of a matrix and a vector. Similarly, channel-wise attention aggregates global information first, and then computes a channel-wise attention weight vector with the hidden state $h_{t-1}$:
\begin{align}
b(h_{t-1}, X) &= \tanh((W_2\text{GAP}(X)+b_2)\oplus W_1h_{t-1}) \\
\Phi_c(h_{t-1}, X) &= \text{Softmax}(W_3(b(h_{t-1}, X))+b_3)    
\end{align}
Overall, the  SCA mechanism can be written in one of two ways. If channel-wise attention is applied before spatial attention, we have
\begin{align}
\label{eq_cs_sca}
Y &= f(X,\Phi_s(h_{t-1}, X \Phi_c(h_{t-1}, X)), \Phi_c(h_{t-1}, X)) 
\end{align}
and  if spatial attention comes first:
\begin{align}
\label{eq_sc_sca}
Y &= f(X,\Phi_s(h_{t-1}, X), \Phi_c(h_{t-1}, X \Phi_s(h_{t-1}, X)))
\end{align}
where $f(\cdot)$ denotes the modulate function which takes the feature map $X$ and attention maps as input and then outputs the modulated feature map $Y$.

Unlike previous attention mechanisms which consider each image region equally and use global spatial information to tell the network where to focus, SCA-Net leverages the semantic vector to produce the spatial attention map as well as the channel-wise attention weight vector. Being more than a powerful attention model, SCA-CNN also provides a better understanding of where and what the model should focus on during sentence generation.

\subsubsection{GALA}


Most attention mechanisms learn where to focus using only weak supervisory signals from class labels, which inspired Linsley et al.~\cite{LinsleySES19} to investigate how explicit human supervision can affect the performance and interpretability of attention models. As a proof of concept, Linsley et al. proposed the \emph{global-and-local attention} (GALA) module, which extends an SE block with a spatial attention mechanism.

Given the input feature map $X$, GALA uses an attention mask that combines global and local attention to tell the network where and on what to focus. As in SE blocks, global attention aggregates global information by global average pooling and then produces a channel-wise attention weight vector using a multilayer perceptron. In local attention, two consecutive $1\times 1$ convolutions are conducted on the input to produce a positional weight map. The outputs of the local and global pathways are combined by addition and multiplication. Formally, GALA can be represented as:
\begin{align}
    s_g &= W_{2} \delta (W_{1}\text{GAP}(x)) \\
    s_l &= \text{Conv}_2^{1\times 1} (\delta(\text{Conv}_1^{1\times1}(X))) \\
    s_g^* &= \text{Expand}(s_g) \\
    s_l^* &= \text{Expand}(s_l) \\
    s &= \tanh(a  (s_g^*+s_l^*)+m\cdot (s_g^* s_l^*)) \\
    Y &= sX
\end{align}
where $a,m \in \mathbb{R}^{C}$ are learnable parameters representing channel-wise weight vectors. 

Supervised by human-provided feature importance maps, GALA has significantly improved representational power and can be combined with any CNN backbone.

\subsection{Spatial \& Temporal Attention}

Spatial \& temporal attention
combines the advantages of spatial attention and temporal attention as it  adaptively selects both important regions and key frames.
Some  works~\cite{song2016end,rstan} compute
temporal attention and spatial attention separately,
while others~\cite{Fu_2019_STA}
produce joint spatiotemporal attention maps. 
Further works focusing on capturing pair-wise relations~\cite{yang2020spatial}.
Representative spatial \& temporal attention attentions 
and specific process $g(x)$ and $f(g(x), x)$ described as Eq.~\ref{eq_general}
are summarised in Tab.~\ref{spatial_temporal_attentions}. We next discuss specific spatial \& temporal attention mechanisms 
\gmh{according to the order in Fig.~\ref{fig:total_attention_summary}}.

\begin{table*}
	\centering
	\caption{Representative spatial \& temporal attentions sorted by date. Action=action recognition, ReID = re-identification. Ranges means the ranges of attention map. S or H means soft or hard attention. $g(x)$ and $f(g(x), x)$ are the attention process described by Eq.~\ref{eq_general}. (A)element-wise product.(B) aggregate information via attention map.(\uppercase\expandafter{\romannumeral1}) emphasize key points in both  spatial and temporal domains, (\uppercase\expandafter{\romannumeral2})capture global information.}
\label{spatial_temporal_attentions}
\begin{tabular}{|p{2cm}|p{2cm}|p{1.5cm}|p{1cm}|p{3cm}|p{1.4cm}|p{1.2cm}|p{0.8cm}|p{1.0cm}|}
\hline
\textbf{Category} & \textbf{Method} & \textbf{Publication} & \textbf{Tasks} & \textbf{$\bm{g(x)}$} & \textbf{$\bm{f(g(x), x)}$} & \textbf{Ranges}  & \textbf{S or H} & \textbf{Goals} \\
\hline
Separately predict spatial \& temporal attention & STA-LSTM~\cite{song2016end} & AAAI2017 & Action &{a)spatial: fuse hidden state -> MLP -> Softmax, b)temporal: fuse hidden state -> MLP -> ReLU} & (A) & {(0,1), (0,+$\infty$)} & {S} & {(\uppercase\expandafter{\romannumeral1}) } \\ 
\cline{2-9}
 & RSTAN~\cite{rstan} & TIP2018 & Action &{a)spatial: fuse hidden state -> MLP -> Softmax, b)temporal: fuse hidden state -> MLP -> Softmax } &(B)&{(0,1)}&{S}&{(\uppercase\expandafter{\romannumeral1}) (\uppercase\expandafter{\romannumeral2})} \\
\hline
Jointly predict spatial \& temporal attention & STA~\cite{Fu_2019_STA} & AAAI2019 & ReID &{a) tenporal: produce per-frame attention maps using $l_2$ norm b) spatial: obtain spatial scores for each patch by summation using $l_1$ norm.} &{(B)}&{(0,1)}&{S} & {(\uppercase\expandafter{\romannumeral1})} \\ 
\hline
Pairwise relation-based method & STGCN~\cite{yang2020spatial} & CVPR2020 & ReID &{ construct a patch graph using pairwise similarity} &{(B)}&{(0,1)}&{S} & {(\uppercase\expandafter{\romannumeral1})} \\ 
\hline
\end{tabular}
\end{table*}

\subsubsection{STA-LSTM}
In human action recognition, 
each type of action  generally only depends 
on a few specific kinematic joints~\cite{song2016end}.
Furthermore, over time, multiple actions may be performed.
Motivated by these observations, Song et al.~\cite{song2016end} proposed 
a joint spatial and temporal attention network based on LSTM~\cite{HochSchm97_lstm}, to adaptively find discriminative features and keyframes. 
Its main attention-related components are a spatial attention sub-network, to select important regions, and a temporal attention sub-network, to select key frames. The spatial attention sub-network can be written as:
\begin{align}
    s_{t} &= U_{s}\tanh(W_{xs}X_{t} + W_{hs}h_{t-1}^{s} + b_{si}) + b_{so}  \\ 
    \alpha_{t} &= \text{Softmax}(s_{t}) \\
    Y_{t} &= \alpha_{t}  X_{t} 
\end{align}
where $X_{t}$ is the input feature at time $t$, $U_{s}$, $W_{hs}$, $b_{si}$, and $b_{so}$ are learnable parameters, and $h_{t-1}^{s}$ is the hidden state at step $t-1$. Note that use of the hidden state $h$ means  the attention process takes  temporal relationships into consideration.

The temporal attention sub-network is similar to the spatial branch and produces its attention map using:
\begin{align}
    \beta_{t} = \delta(W_{xp}X_{t} + W_{hp}h_{t-1}^{p} + b_{p}). 
\end{align}
It adopts a ReLU function instead of a normalization function for ease of optimization. It also uses a regularized objective function to improve  convergence.

Overall, this paper presents a joint spatiotemporal attention method
to focus on important joints and keyframes, 
with excellent results on the action recognition task.

\subsubsection{RSTAN}
To capture  spatiotemporal contexts in video frames,
Du et al.~\cite{rstan} introduced \emph{spatiotemporal attention}
to adaptively identify key features in a global way. 

The spatiotemporal attention mechanism in RSTAN consists of a spatial attention module and a temporal attention module applied serially. Given an input feature map $X\in \mathbb{R}^{D\times T\times H\times W}$ and the previous hidden state $h_{t-1}$ of an RNN model, spatiotemporal attention aims to produce a spatiotemporal feature representation for action recognition. First, the given feature map $X$ is reshaped to $\mathbb{R}^{D\times T\times (H\times W)}$, and we define $X(n,k)$ as the feature vector for the $k$-th location of the $n$-th frame. At time $t$, the spatial attention mechanism aims to produce a global feature $l_n$ for each frame, which can be written as
\begin{align}
    \alpha_t(n,k) &= w_\alpha  \tanh(W_hh_{t-1}+W_{x}X(n,k)+b_\alpha) \\
    \alpha_t^*(n,k) &= {e^{\gamma_\alpha\alpha_t(n,k)}}/{\sum_{j=1}^{W\times H}e^{\gamma_\alpha\alpha_t(n,k)}} \\
    l_n &= \sum_{k=1}^{H\times W}\alpha_t^*(n,k) X(n,k) 
\end{align}
where $\gamma_\alpha$ is introduced to control the sharpness of the location-score map. After obtaining frame-wise features $\{l_1,\dots,l_T\}$, RSTAN uses a temporal attention mechanism to estimate the importance of each frame feature
\begin{align}
    \beta_t(n) &= w_\beta  \tanh(W_h'h_{t-1}+W_{l}l(n)+b_\beta) \\
    \beta_t^*(n) &= {e^{\gamma_\beta\beta_t(n)}}/{\sum_{j=1}^{T}e^{\gamma_\beta\beta_t(n)}} \\
    \phi_t &= \sum_{n=1}^{T}\beta_t^*(n) l(n)
\end{align}

The spatiotemporal attention mechanism used in RSTAN  identifies those regions in both spatial and temporal domains which are strongly related to the prediction in the current step of the RNN. This efficiently enhances the representation power of any 2D CNN.

\subsubsection{STA}

Previous attention-based methods for video-based person re-identification only assigned an attention weight to each frame and failed to capture joint spatial and temporal relationships. To address this issue, Fu et al.~\cite{Fu_2019_STA} propose a novel \emph{spatiotemporal attention} (STA) approach, which assigns attention scores for each spatial region in different frames without any extra parameters. 

Given the feature maps of an input video $\{X_n|X_n\in \mathbb R^{C\times H\times W}\}_{n=1}^N$, STA first generates frame-wise attention maps by using the $l_2$ norm on the squares sum in the channel domain:
\begin{align}
    g_n(h,w) = \frac{||\sum_{c=1}^C X_n(c,h,w)^2||_2}{\sum_{h=1}^H \sum_{w=1}^W ||\sum_{c=1}^C X_n(c,h,w)^2||_2} 
\end{align}
Then both the feature maps and attention maps are divided into $K$ local regions horizontally, each of which represents one part of the person. The spatial attention score for region $k$ is obtained using 
\begin{align}
    s_{n,k} = \sum_{(i,j)\in \text{Region}_k}||g_n(i,j)||_1
\end{align}
To capture the relationships between regions in different frames, STA applies  $l_1$ normalization to the attention scores in the temporal domain, using
\begin{align}
    S(n,k) = \frac{s_{n,k}}{\sum_{n=1}^N ||s_{n,k}||_1}
\end{align}
Finally, STA splits the input feature map $X_i$ into $K$ regions $\{X_{n,1},\dots,X_{n,K}\}$ and computes the output  using
\begin{align}
    Y^1 &= [X_{\arg\max_{n}S(n,1),1};\dots;X_{\arg\max_{n}S(n,K),K}] \\
    Y^2 &=  [\sum_{n=1}^N S(n,1)X_{n,1};\dots;\sum_{n=1}^N S(n,K)X_{n,K}] \\
    Y &= [Y^1;Y^2]
\end{align}

Instead of computing spatial attention maps frame by frame, STA considers  spatial and temporal attention information simultaneously, fully using the discriminative parts in both dimensions. This reduces the influence of occlusion. Because of its non-parametric design, STA can tackle input video sequences of variable length; it can be combined with any 2D CNN backbone.   

\subsubsection{STGCN}

To model the spatial relations within a frame and temporal relations across frames, Yang et al.~\cite{yang2020spatial} proposed a novel \emph{spatiotemporal graph convolutional network} (STGCN) to learn a discriminative descriptor for a video. It constructs a patch graph using pairwise similarity, and then uses graph convolution to aggregate information.

STGCN includes two parallel GCN branches, the temporal graph module and the structural graph module. Given the feature maps of a video, STGCN first horizontally partitions each frame into $P$ patches and applies average pooling to generate patch-wise features ${x_1,\dots,x_N}$, where the total number of patches is $N=T P$. For the temporal module, it takes each patch as a graph node and construct a patch graph for the video, where the adjacency matrix $\widehat{A}$ is obtained by  normalizing the pairwise relation matrix $E$, defined as
\begin{align}
    E(i,j) &= (W^\phi x_i)^T W^\phi x_j \\
    A(i,j) &= {E^2(i,j)}/{\sum_{j=1}^N E^2(i,j)} \\
    \widehat{A} &= D^{-\frac{1}{2}}(A+I)D^{-\frac{1}{2}}
\end{align}
where $D(i,i) = \sum_{j=1}^N (A+I)(i,j)$. Given the adjacency matrix $\widehat{A}$, the $m$-th graph convolution can be found using 
\begin{align}
    X^m = \widehat{A} X^{m-1} W^m + X^{m-1}
\end{align}
where $X \in \mathbb{R}^{N \times c}$ represents the hidden features for all patches and $W^m \in \mathbb{R}^{c\times c}$ denotes the learnable weight matrix for the $m$-th layer. For the spatial module, STGCN follows a similar approach of adjacency matrix and graph convolution, except for modeling the spatial relations of different regions within a frame.

Flattening spatial and temporal dimensions into a sequence, STGCN applies the GCN to capture the spatiotemporal relationships of patches across  different frames. Pairwise attention is used to obtain the weighted adjacency matrix. By leveraging spatial and temporal relationships between patches, STGCN overcomes the occlusion problem while also enhancing informative features. It can used with any CNN backbone to process  video.

\section{Future directions} \label{sec:direction}
We present our thoughts on potential future research directions.

\subsection{\gmh{Necessary and sufficient condition for attention}}
We find the Eq.~\ref{eq_general} is a necessary condition but not a necessary and sufficient condition. 
For instance, GoogleNet~\cite{szegedy2014going} conforms to the above formula, but does not belong to the attention mechanisms.
Unfortunately, we find it difficult to find a necessary and sufficient condition for all attention mechanisms. 
The necessary and sufficient conditions for the attention mechanism are still worth exploring which can promote our understanding of attention mechanisms.

\subsection{General attention block}
At present, a special attention mechanism needs to be designed for each different task, which requires considerable effort to explore potential attention methods. For instance, channel attention is a good choice for image classification, while 
\gmh{spatial attention} 
is well-suited to dense prediction tasks such as semantic segmentation and object detection. Channel attention focuses on \emph{what to pay attention to} while spatial attention considers  \emph{where to pay attention}. Based on this observation, we encourage consideration as to whether there could be a general attention block that takes advantage of all kinds of attention mechanisms. For example,  a soft selection mechanism (branch attention) could choose between channel attention, spatial attention and temporal attention according to the specific task undertaken.

\subsection{Characterisation and interpretability}
Attention mechanisms are motivated by the human visual system and are a step towards the goal of building an interpretable computer vision system. Typically, attention-based models are understood by rendering attention maps, as in Fig.~\ref{fig:attention_map_vit}. However, this can only give  an intuitive feel for what is happening, rather than precise understanding. However, applications in which security or safety are important, such as  medical diagnostics and automated driving systems, often have stricter requirements. Better characterisation of how methods work, including modes of failure, is needed in such areas. Developing characterisable and  interpretable attention models could make them  more widely applicable.

\subsection{\gmh{Sparse activation}}  
We visualize some attention map and obtains consistent conclusion with ViT~\cite{dosovitskiy2020vit} shown in Fig.~\ref{fig:attention_map_vit} that attention mechanisms can produce sparse activation. There phenomenon give us a inspiration that sparse activation can achieve a strong performance in deep neural networks. It is worth noting that sparse activation is similar with human cognition. Those motivate us to explore which kind of architecture can simulate human visual system.

\subsection{Attention-based pre-trained models}

Large-scale attention-based pre-trained models have had great success in natural language processing~\cite{brown2020language, bao2021beit}. Recently, MoCoV3~\cite{chen2021empirical_mocov3}, DINO~\cite{caron2021emerging},  BEiT~\cite{bao2021beit} and MAE~\cite{he2021masked} have demonstrated that attention-based models are also well suited to  visual tasks. Due to their ability to adapt to varying inputs, attention-based models can deal with unseen objects and are naturally suited to transferring  pretrained weights to a variety of tasks. We believe that the combination of pre-training and attention models should be further explored: training approach, model structures, pre-training tasks and the scale of data are all worth investigating.


\subsection{Optimization}
SGD~\cite{qian99_sgd} and Adam~\cite{kingma2017adam} are well-suited for optimizing convolutional neural networks. For visual transformers, AdamW~\cite{loshchilov2019decoupled} works better. Recently, Chen et al.~\cite{chen2021vision} significantly improved visual transformers by using a novel optimizer, the  \emph{sharpness-aware minimizer} (SAM)~\cite{foret2021sharpnessaware}. It is clear that attention-based networks and convolutional neural networks are different models; different optimization methods may work better for different models. Investigating new optimzation methods for attention models is likely to be worthwhile.

\subsection{Deployment}
Convolutional neural networks have a simple, uniform structure which makes them easy to deploy on various hardware devices. However, it is difficult to optimize complex and varied attention-based models on edge devices.  Nevertheless, experiments in~\cite{liu2021swin, wu2021cvt, yuan2021volo} show that attention-based models provide better results than convolutional neural networks, so it is worth trying to find simple, efficient and effective attention-based models which can be widely deployed.

\section{Conclusions} \label{sec:conclusion}
Attention mechanisms have become an indispensable technique in the field of computer vision in the era of deep learning. 
This survey has systematically reviewed and summarized attention mechanisms for deep neural networks in computer vision. 
We have grouped different attention methods according to their domain of operation, rather than by application task, and show that
 attention models can be regarded as an independent topic
in their own right. 
We have concluded  with some potential directions for future research.
We  hope that this work will encourage a variety of potential application developers to put attention mechanisms to use to improve their deep learning results.
We also hope that this survey will give researchers 
a deeper understanding of various attention mechanisms and the relationships between them, as a springboard for future research.






%

\ifCLASSOPTIONcompsoc
  \section*{Acknowledgments}
  This work was supported by the Natural Science Foundation of China (Project 61521002, 62132012). We would like to thank Cheng-Ze Lu, Zhengyang Geng, Shilong liu, He Wang, Huiying Lu and Chenxi Huang for their helpful discussions and insightful suggestions.
\else
  \section*{Acknowledgment}
\fi


\ifCLASSOPTIONcaptionsoff
  \newpage
\fi

\bibliographystyle{IEEEtran}
\bibliography{egbib}

\end{document}